\documentclass[conference]{IEEEtran}
\IEEEoverridecommandlockouts
\usepackage{comment}
\usepackage{cite}
\usepackage{amsmath,amssymb,amsfonts}
\usepackage{algorithmic}
\usepackage[ruled,vlined]{algorithm2e}
\usepackage{graphicx}
\usepackage{textcomp}
\usepackage{xcolor}
\usepackage{amsmath}
\usepackage{caption}
\usepackage{subcaption}
\usepackage[normalem]{ulem}
\usepackage{soul}
\usepackage{booktabs}
\usepackage{multicol}
\usepackage{pdfpages}
\usepackage{multirow}

\def\BibTeX{{\rm B\kern-.05em{\sc i\kern-.025em b}\kern-.08em
    T\kern-.1667em\lower.7ex\hbox{E}\kern-.125emX}}
    
\usepackage{fancyhdr}
\usepackage{kantlipsum}
\usepackage{url}

\usepackage[normalem]{ulem}
\usepackage{bm}
\usepackage{dsfont}
\newcommand{\bx}{{\bm x}}
\newcommand{\Real}{\mathds{R}} %

\newcommand{\cpg}[1]{{\color{blue}{#1}}}

\fancypagestyle{plain}{
\fancyhf{}
\fancyhead[C]{Conference on \LaTeX} 

}
\usepackage{eso-pic}
\newcommand{\fixme}[1]{{\color{red}\em\bf{[FIXME: #1]}}}

\newcommand{\shakshi}[1]{{\color{magenta} [{\textbf{Shakshi: }}\color{magenta}{#1}}{\color{magenta} ]}}
\begin{document}

\AddToShipoutPictureBG*{
\AtPageUpperLeft{

\setlength\unitlength{1in}

\hspace*{\dimexpr0.5\paperwidth\relax}

}
}

\title{Misinformation Detection on YouTube Using Video Captions}

\author{
\IEEEauthorblockN{Raj Jagtap}
\IEEEauthorblockA{
\textit{School of Mathematics and Computer Science}\\
Indian Institute of Technology Goa, India \\
\texttt{raj.jagtap.17003@iitgoa.ac.in}}
\and
\IEEEauthorblockN{Abhinav Kumar}
\IEEEauthorblockA{\textit{School of Mathematics and Computer Science} \\
Indian Institute of Technology Goa, India\\
\texttt{abhinav.kumar.17001@iitgoa.ac.in}}
\and
\IEEEauthorblockN{Rahul Goel}
\IEEEauthorblockA{\textit{Institute of Computer Science} \\
University of Tartu, Tartu, Estonia\\
\texttt{rahul.goel@ut.ee}}
\and
\IEEEauthorblockN{Shakshi Sharma}
\IEEEauthorblockA{\textit{Institute of Computer Science} \\
University of Tartu, Tartu, Estonia\\
\texttt{shakshi.sharma@ut.ee}}\\
\and
\IEEEauthorblockN{Rajesh Sharma}
\IEEEauthorblockA{\textit{Institute of Computer Science} \\
University of Tartu, Tartu, Estonia\\
\texttt{rajesh.sharma@ut.ee}}
\and
\IEEEauthorblockN{Clint P. George}
\IEEEauthorblockA{\textit{School of Mathematics and Computer Science} \\
Indian Institute of Technology Goa, India\\
\texttt{clint@iitgoa.ac.in}}

}

\maketitle
\IEEEoverridecommandlockouts
\IEEEpubidadjcol

\begin{abstract}  
Millions of people use platforms such as YouTube, Facebook, Twitter, and other mass media. Due to the accessibility of these platforms, they are often used to establish a narrative, conduct propaganda, and disseminate misinformation. This work proposes an approach that uses state-of-the-art NLP techniques to extract features from video captions (subtitles). To evaluate our approach, we utilize a publicly accessible and labeled dataset for classifying videos as misinformation or not. The motivation behind exploring video captions stems from our analysis of videos metadata. Attributes such as the number of views, likes, dislikes, and comments are ineffective as videos are hard to differentiate using this information. Using caption dataset, the proposed models can classify videos among three classes (Misinformation, Debunking Misinformation, and Neutral) with $0.85 \text{ to } 0.90$ F$1$-score. To emphasize the relevance of the misinformation class, we re-formulate our classification problem as a two-class classification - Misinformation vs. others (Debunking Misinformation and Neutral). In our experiments, the proposed models can classify videos with $0.92 \text{ to } 0.95$ F$1$-score and $0.78 \text{ to } 0.90$ AUC ROC.
\end{abstract}


\section{Introduction and Overview}\label{sec:intro}

Online Social Media (OSM) platforms has ushered in a new era of ``misinformation'' by disseminating \textit{incorrect} or \textit{misleading information} to deceive users \cite{guess2019less}. OSM platforms, such as Twitter, Facebook, YouTube, which initially became popular due to their social aspect (connecting users), have become alternative platforms for sharing and consuming news \cite{lee2017effects}. Due to no third-party verification of content, the users on these platforms often (consciously or unconsciously) engage in the spreading of misinformation or Fake News \cite{ShakshiIJCNN2021,bovet2019influence,lemos2020fake}. 
Especially, YouTube is one of the most popular OSM platform, is ideal for injecting misinformation. 



According to recent research, YouTube, the largest video sharing platform with a user base of more than 2 billion users, is commonly utilized to disseminate misinformation and hatred videos \cite{covid-misinfo}. According to a survey \cite{clement2019hours}, $74\%$ of adults in the USA use YouTube, and approximately 500 hours of videos are uploaded to this platform every minute, which makes YouTube hard to monitor. Thus, this makes YouTube an excellent forum for injecting misinformation videos, which could be difficult to detect among the avalanche of content \cite{donzelli2018misinformation,goobie2019youtube}. This has generated severe issues, necessitating creative ways to prevent the spread of misinformation on this platform.

In this work, we propose an approach for detecting misinformation among YouTube videos by utilizing an existing YouTube dataset \cite{YTdataset}, containing metadata such as title, description, views, likes, dislikes. The dataset covers videos related to five different topics, namely, (1) Vaccines Controversy, (2) 9/11 Conspiracy, (3) Chem-trail Conspiracy, (4) Moon Landing Conspiracy, and (5) Flat Earth Theory. Each video is labeled from one of the three categories, that is, i) misinformation, ii) debunking misinformation and iii) neutral (not related to misinformation). In total, this off-the-shelf dataset contains 2943 unique videos. In addition, to existing metadata, we also downloaded captions of the videos for detecting videos related to misinformation.

The motivation behind analyzing captions is inspired from the results of our descriptive analysis, where we observe that basic meta-information about Youtube videos, such as the number of views, likes, dislikes, are similar, especially between misinformation and debunking misinformation categories (details in Section \ref{Descriptive Analysis}). In addition, the titles and descriptions may mislead in some cases. For example, videos with titles \emph{Question \& Answer at The 2018 Sydney Vaccination Conference}\footnote{https://www.youtube.com/watch?v=I7gZGOrPPv0}, 
and 
\emph{The Leon Show - Vaccines and our Children}\footnote{https://www.youtube.com/watch?v=AFnh06drH48} do not indicate any misinformation content, although the videos indeed communicate misinformation. Precisely, the former video conveys that good proteins and vitamins help one live a healthier life than vaccines, and most of the immunity is in the gut that vaccines can destroy. In the latter video, a physician describes some information scientifically; however, in between, the person indicates autism rates have skyrocketed after vaccines came into existence. 

In this work, we build multi-class prediction models for 
classifying videos into three classes: Misinformation, Debunking Misinformation, and  Neutral  using various Natural Language Processing (NLP) approaches. 
In our experiments the proposed models can classify videos among three classes with $0.85 \text{ to } 0.90$ F$1$-score. Next, to emphasize the misinformation class's relevance, we re-formulate our classification problem as a two-class classification - Misinformation vs. the other two (Debunking Misinformation, and Neutral). The proposed model (s) can classify videos among two classes with $0.92 \text{ to } 0.95$ F$1$-score and $0.78 \text{ to } 0.90$ AUC ROC.

The rest of the paper is organized as follows. Next, we discuss the related work. We then describe the dataset in Section \ref{Sec:Dataset}. Section \ref{sec:methodology} presents our methodology and various machine learning models to classify videos into different categories is covered in Section \ref{sec:experiments}. We conclude with a discussion of future directions in Section \ref{Sec:Conclusion}.

\section{Related Work}\label{sec:related-work}


Misinformation on online platforms can greatly impact human life. For instance, spreading of a large number of misinformation news during the US presidential election of 2016 scared the world about the possible worldwide impact of misinformation. In the past, researchers have studied misinformation on OSM platforms such as Facebook \cite{allcott2017social,guess2019less}, Twitter \cite{bovet2019influence,ShakshiIJCNN2021}, Instagram \cite{mena2020misinformation}, YouTube \cite{hussain2018analyzing,lemos2020fake,donzelli2018misinformation} covering various topics such as politics \cite{allcott2017social,bovet2019influence,lemos2020fake}, healthcare \cite{bautista2021healthcare}, and performing tasks such as empirical analysis of Fake News articles \cite{allcott2017social}, identifying characteristics of individuals involved in spreading of Fake News \cite{guess2019less}, and predicting rumor spreaders \cite{ShakshiIJCNN2021} to name a few.

Even though all significant OSM platforms can contain misinformation, research indicates that YouTube has played an especially crucial role as a source of misinformation \cite{jiang2019bias,carne2019conspiracies,diresta2018complexity,weissman2019despite}. In \cite{YTdataset}, the authors investigated whether personalization (based on gender, age, geolocation, or watch history) contributes to spread the misinformation on the YouTube search engine. They conclude that gender, age, or geolocation do not significantly amplify misinformation in returned search results for users with new accounts. However, once the user develops some watch history, the personal attributes start playing a significant role in recommending misinformation to the users.

Past research has also thoroughly examined the capabilities and limitations of NLP for identifying misinformation. Theoretical frameworks for analyzing the linguistic and contextual characteristics of many forms of misinformation, including rumors, fake news, and propaganda, have been developed by researchers \cite{li-etal-2019-rumor-detection,thorne-vlachos-2018-automated,rubin2016fake,zhou2018fake}. Given the difficulties in identifying misinformation in general, researchers have also created specialized benchmark datasets to assess the performance of NLP architectures in misinformation-related classification tasks \cite{perez-rosas-etal-2018-automatic,hanselowski-etal-2018-retrospective}. Various research papers suggest case-specific NLP techniques for tracing misinformation in online social networks, due to the appearance of a large volume of misinformation. 

In \cite{della2018automatic} and \cite{popat-etal-2018-declare}, the authors integrated linguistic characteristics of articles with additional meta-data to detect fake news. In another line of works\cite{volkova2017separating,qazvinian2011rumor,kumar-carley-2019-tree}, the authors developed special architectures that take into account the microblogging structure of online social networks, while \cite{de-sarkar-etal-2018-attending,gupta-etal-2019-neural} used sentence-level semantics to identify misinformation. Despite the adoption of such fact-checking systems, identifying harmful misinformation and deleting it quickly from social media such as YouTube remains a difficult task \cite{gillespie2018custodians,roberts2019behind}.

According to YouTube's ``Community Guidelines'' enforcement report, the company deletes videos that breach its standards on a large scale. Between January and March $2021$, more than $9$ million videos were deleted due to a violation of the Community Guidelines. The bulk of them was taken down owing to automatic video flagging, with YouTube estimating that 67\% of these videos were taken down before they hit $10$ views \cite{GoogleReport}. However, in \cite{li2020youtube,frenkel2020plandemic,ferrara2020types}, the authors argue that a substantial percentage of conspiratorial information remains online on YouTube and other platforms, influencing the public despite so much research. Given this, it is essential to investigate the misinformation and how we can manage it accordingly. Towards this, we illustrate how NLP-based feature extraction \cite{shu2017fake,jiang2020modeling} based on videos caption can be effectively used for this task. Previous studies explicitly employed comments as proxies for video content classification \cite{momeni2013properties,huang2010text,filippova2011improved,eickhoff2013exploiting,dougruoz2017text}. However, in this work, we analyzed videos caption for classifying videos into Misinformation, and non-misinformation classes (Misinformation-Debunking, and Neutral).





\section{Dataset}\label{Sec:Dataset}
This section first describes the YouTube videos dataset and the Caption Scraper we created to collect the additional data, namely the captions of the videos. Next, we explain the caption pre-processing and some descriptive analysis in the following subsections.


\subsection{YouTube Videos Dataset}\label{sec:YTVD}

For this study, we have used the existing dataset published by \cite{YTdataset}. The dataset was collected over a time of several months for the year $2020$ using an automated approach and covers five different topics, namely, (1) Vaccines Controversy, (2) 9/11 Conspiracy, (3) Chem-trail Conspiracy, (4) Moon Landing Conspiracy, and (5) Flat Earth Theory. For each topic, specific query strings were used to extract the videos from the YouTube search engine using different Google accounts. In total, the original dataset contains $2943$ unique videos (refer Table \ref{table:VideosTopicDistribution}, Column 2 for count of videos in each topic) collected over the topics. Each video is labeled one of the three classes, namely i) Misinformation, ii) Debunking Misinformation, and iii) Neutral (that is not related to misinformation). We manually surfed through the links, and we found that some of the videos were removed by YouTube due to their misinforming content. Table \ref{table:VideosTopicDistribution}, and Figure \ref{fig:my_label} provides information about the videos that are at the time of this work were available on YouTube and used for our analysis. We used our own built scraper to fetch the pre-available captions and download them as text files for further processing. We also observe a general trend where the count of neutral videos for each topic is more than the misinformation and debunking misinformation classes (see Figure \ref{fig:my_label}). In particular, the Chem-trail Conspiracy topic has approximately three times more misinformation videos than those debunking misinformation.


\subsection{Caption Data Collection From YouTube and Preprocessing}\label{subsec:captionCollPreprocess}

Here, we discuss our publicly available YouTube Caption Scraper. The Caption Scraper takes the YouTube video URL as the input and returns a text file containing captions as the output. It uses \textit{YouTube-Caption-API} to get the subtitles of the video. If the subtitles are not present in English, it translates the available subtitles to English. If the channel does not provide the manual subtitles, it uses YouTube's auto-captions feature to get subtitles.

\begin{figure}
    \centering
    \includegraphics[width=\columnwidth]{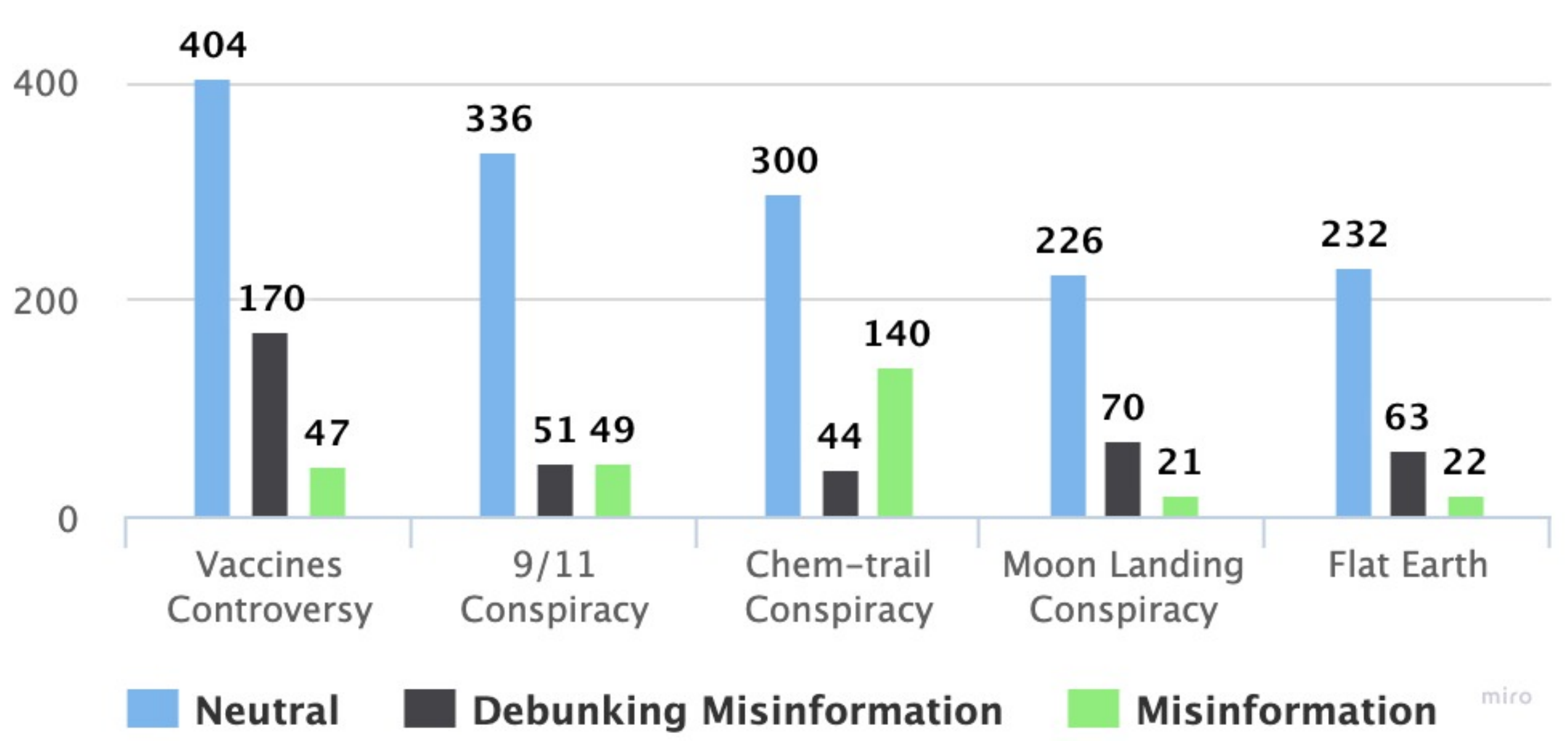}
    \caption{Distribution of classes with respect to each topic.}
    \label{fig:my_label}
\end{figure}

In this work, we used the Caption Scraper script to download captions (subtitles) of the videos present in the dataset. Please note that as YouTube removed some videos and some videos did not provide text captions, we could download captions for only $2175$ videos. The distribution of videos with captions among different topics is shown in Table \ref{table:VideosTopicDistribution}. We also show the distribution of classes with respect to each topic in Figure \ref{fig:my_label}. We can observe that the dataset contains ``Neutral'' class videos in the majority for each topic. This can also be observed that classes are imbalanced among all the topics, which means in all topics, each class has different number of videos.

\begin{table}
\centering
\begin{tabular}{|c||c|p{2cm}|}
 \hline
 Topic&Original Count&Available videos with captions\\
 \hline 
 Vaccines Controversy&$775$&$621$ ($28.5\%$)\\
 9/11 Conspiracy &$654$&$436$ ($20.0\%$)\\
 Chem-trail Conspiracy &$675$&$484$ ($22.2\%$)\\
 Moon Landing Conspiracy &$466$&$317$ ($14.6\%$)\\
 Flat Earth &$373$&$317$ ($14.6\%$)\\
 \hline
\end{tabular}
\caption{Topic-wise distribution of videos}
\label{table:VideosTopicDistribution}
\end{table}

\paragraph{Caption preprocessing} After downloading the captions, we performed multi-stage cleaning (or preprocessing) of them. We start our caption preprocessing by removing all the special symbols (e.g. `!', `?', `.') and numbers. This made the caption text of only alphabets and spaces. 
Furthermore, videos with less than 500 characters in the caption are discarded. This is done to prevent noisy data in caption vectors. For example, we found a few videos with only an introduction of the presenter and then a mute presentation. Now, our final dataset contains 2125 videos in total mentioned in detail in Table \ref{table:datasetStats}, Row 5 (Videos with captions after filtering). Afterwards, using bag of words from each caption document, we removed all the stop-words to prevent caption vectors from being diluted from common, meaningless words. 

\paragraph{Code and caption dataset availability} The source code and caption dataset used in this work is publicly available at \url{https://github.com/jagtapraj123/YT-Misinformation}.

\subsection{Descriptive Analysis}\label{Descriptive Analysis}
This section delves further into the features of YouTube videos, such as the number of views, likes, dislikes, and comments. As previously mentioned, we begin our work with the YouTube dataset released by \cite{YTdataset}, which contains some additional information about the videos apart from the URL, such as the title, the number of views, the number of likes, and dislikes. We explored this additional information to find some trends in the dataset.

\begin{figure*}[ht!]
\begin{tabular}{|c|c|c|c|c|}
\hline
{} & Views & Likes & Dislikes & Comments\\\hline
\rotatebox{90}{Vaccine Controversy}
& 
\subfloat{\includegraphics[width=0.45\columnwidth]{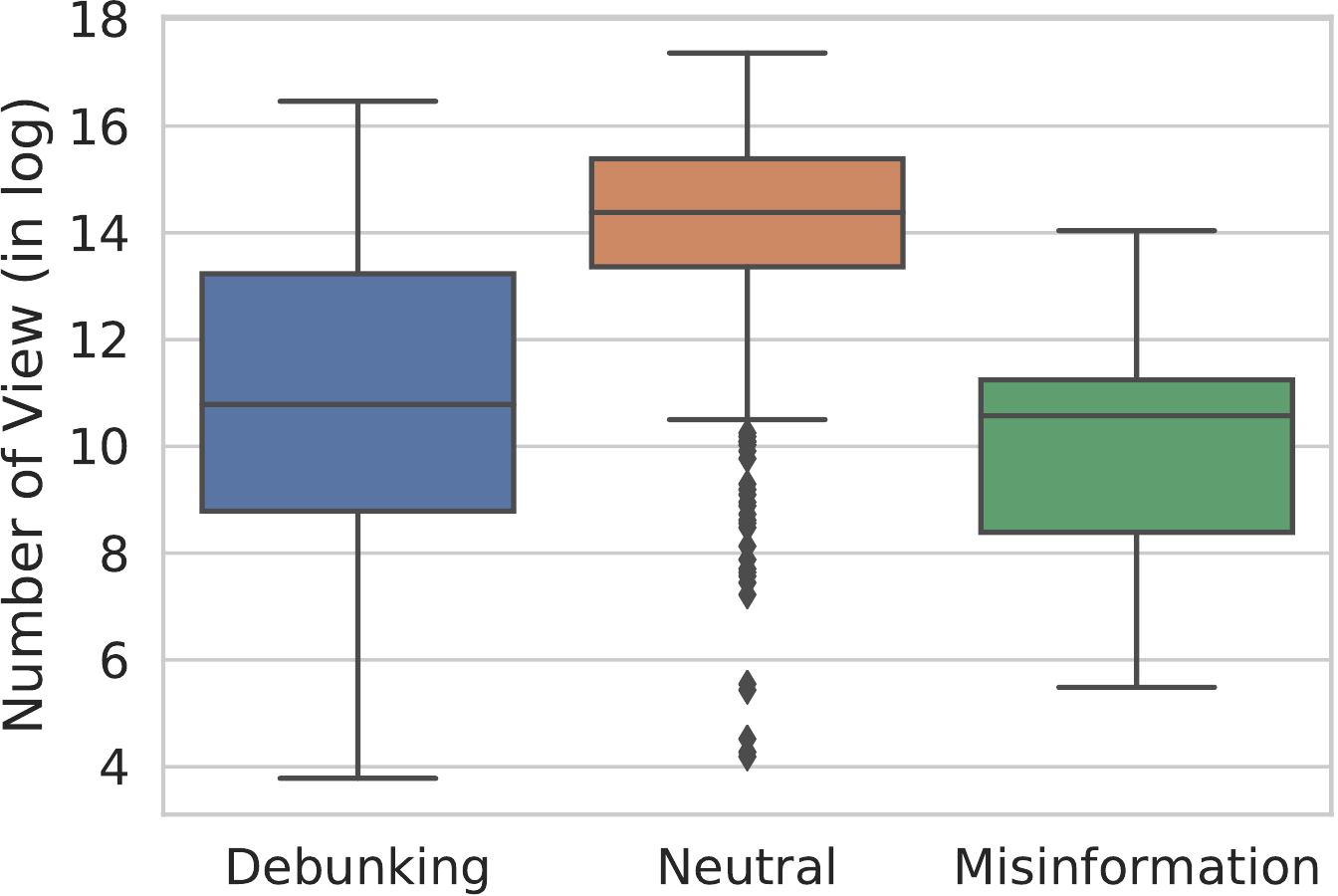}} &
\subfloat{\includegraphics[width=0.45\columnwidth]{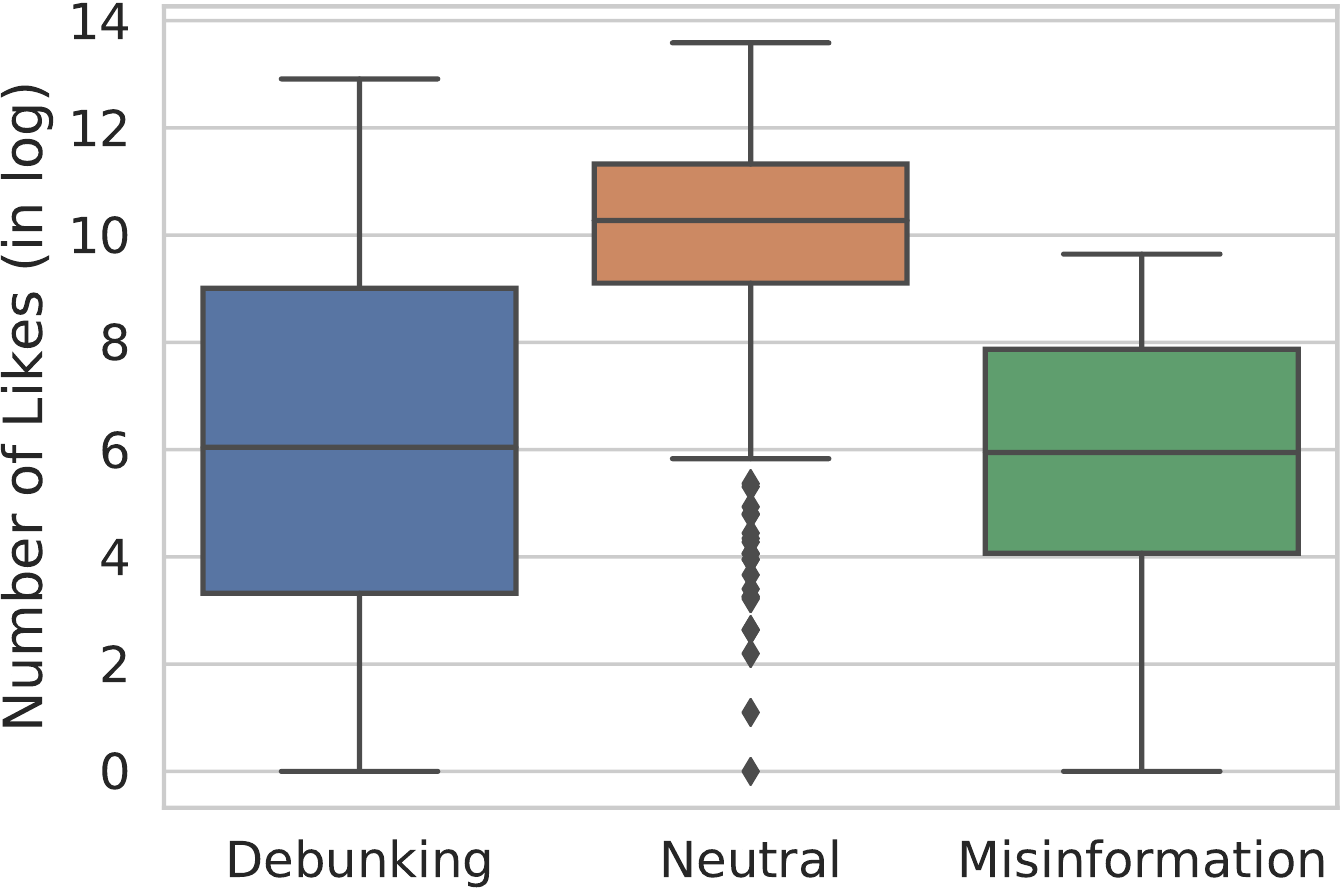}} &
\subfloat{\includegraphics[width=0.45\columnwidth]{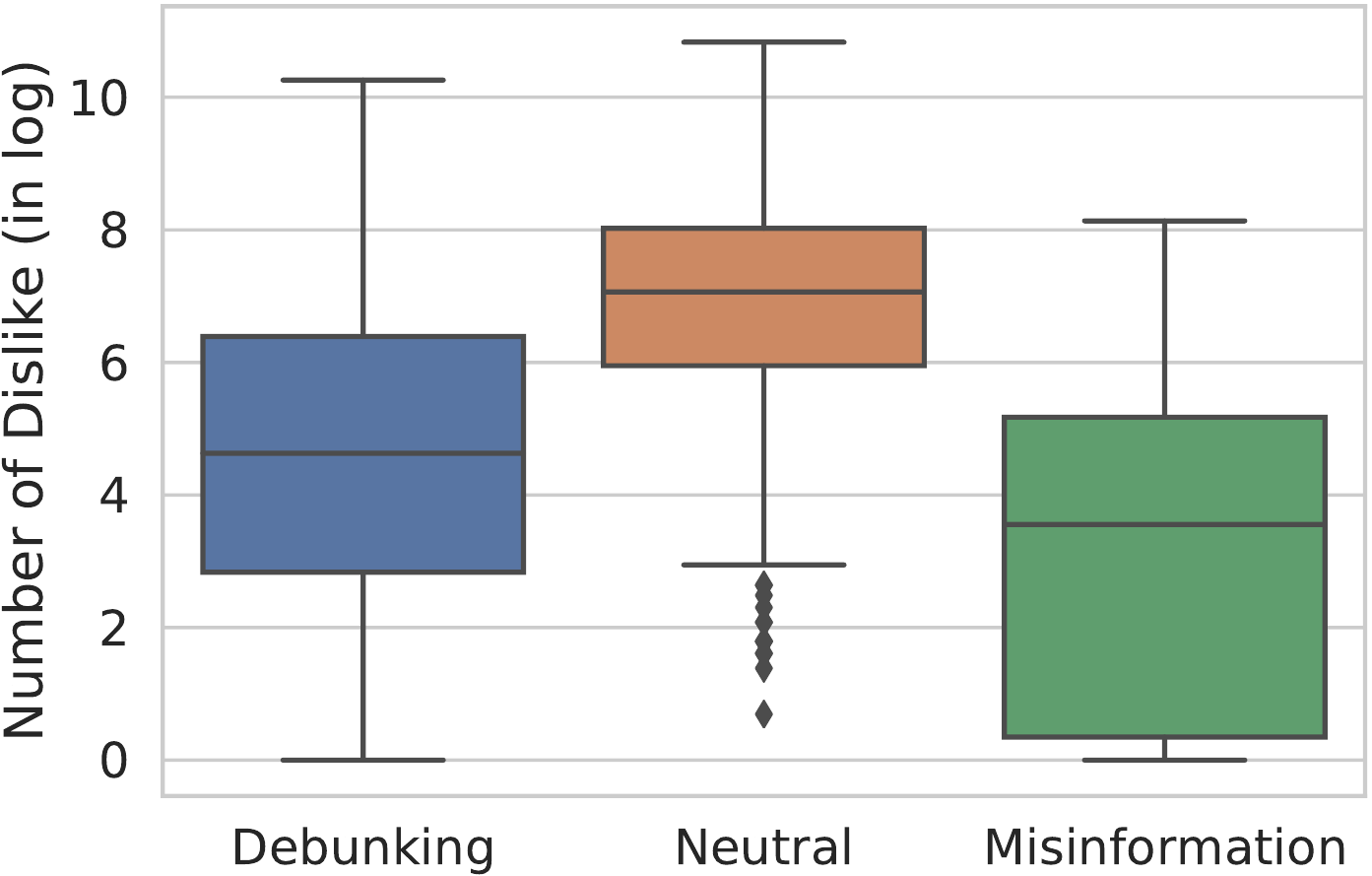}} &
\subfloat{\includegraphics[width=0.45\columnwidth]{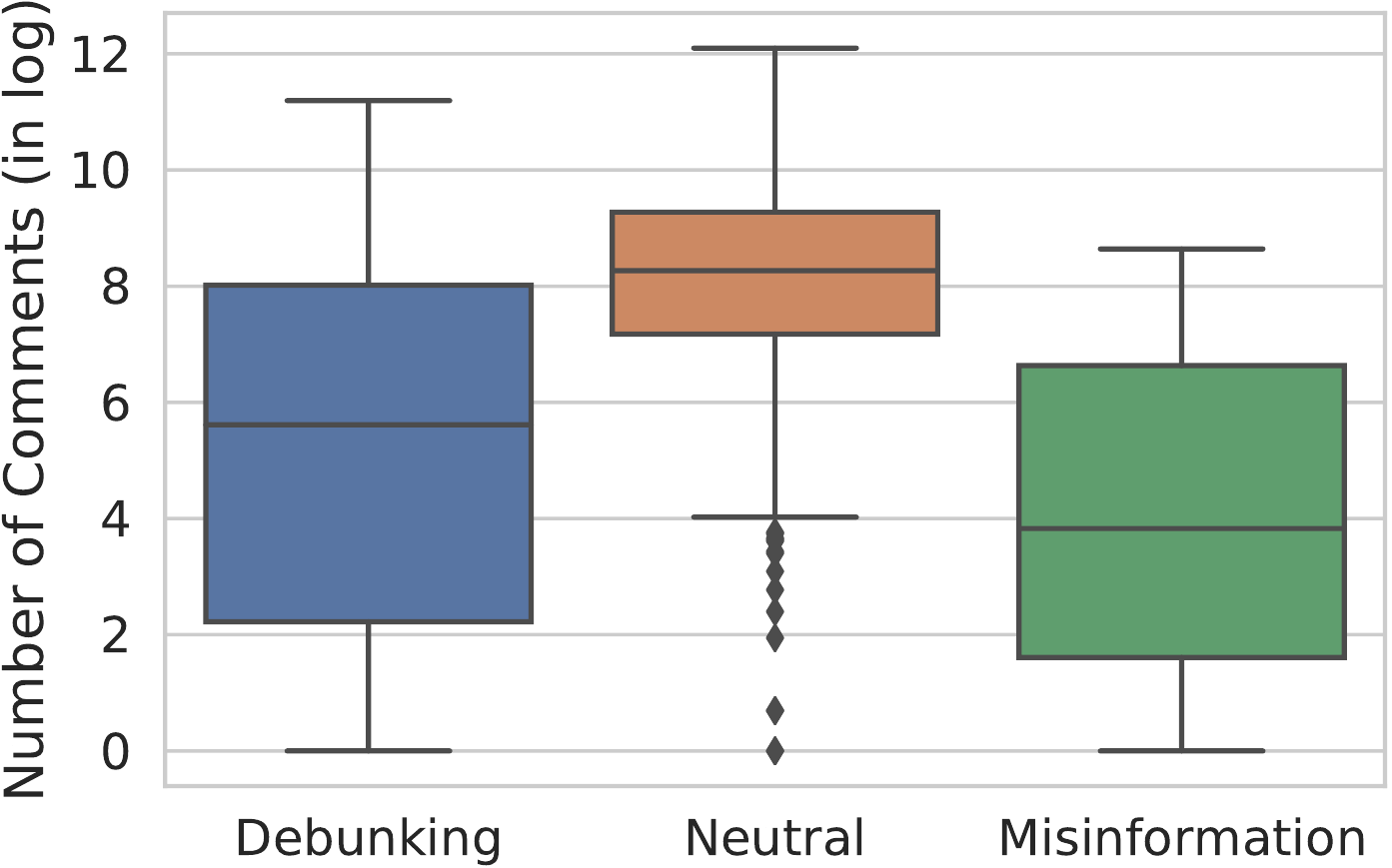}}\\\hline

\rotatebox{90}{9/11 Conspiracy} & 
\subfloat{\includegraphics[width=0.45\columnwidth]{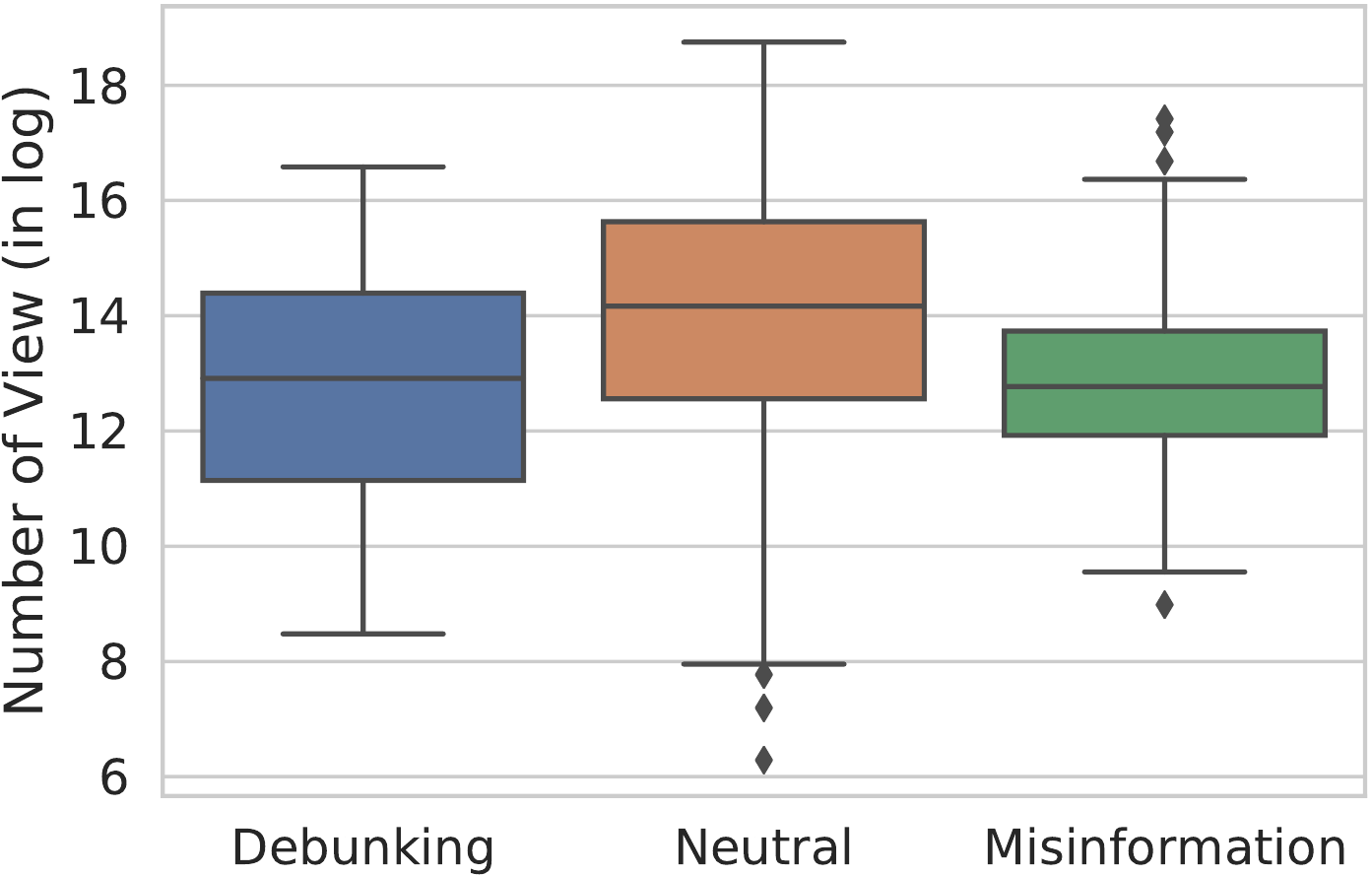}} &
\subfloat{\includegraphics[width=0.45\columnwidth]{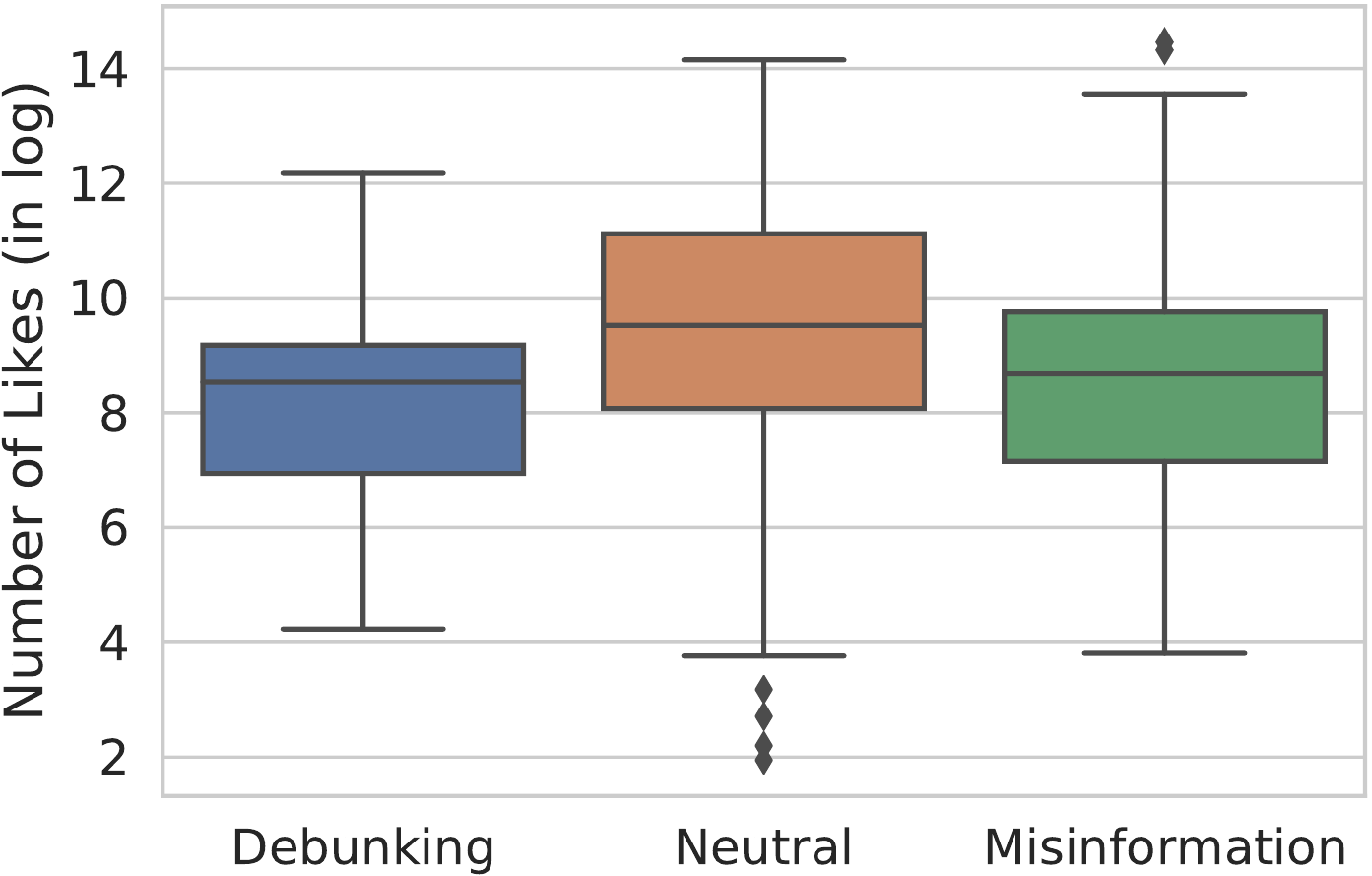}} &
\subfloat{\includegraphics[width=0.45\columnwidth]{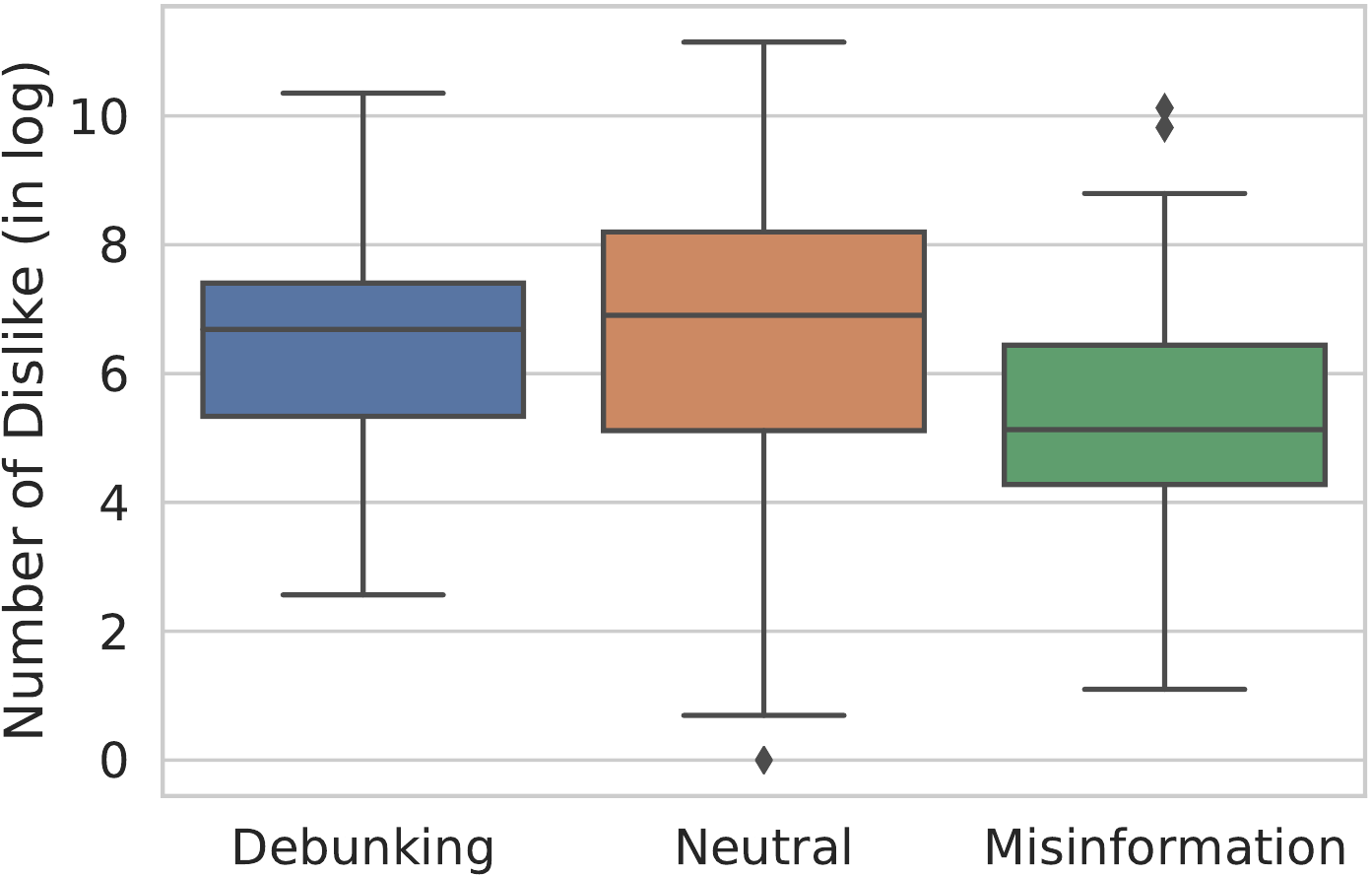}} &
\subfloat{\includegraphics[width=0.45\columnwidth]{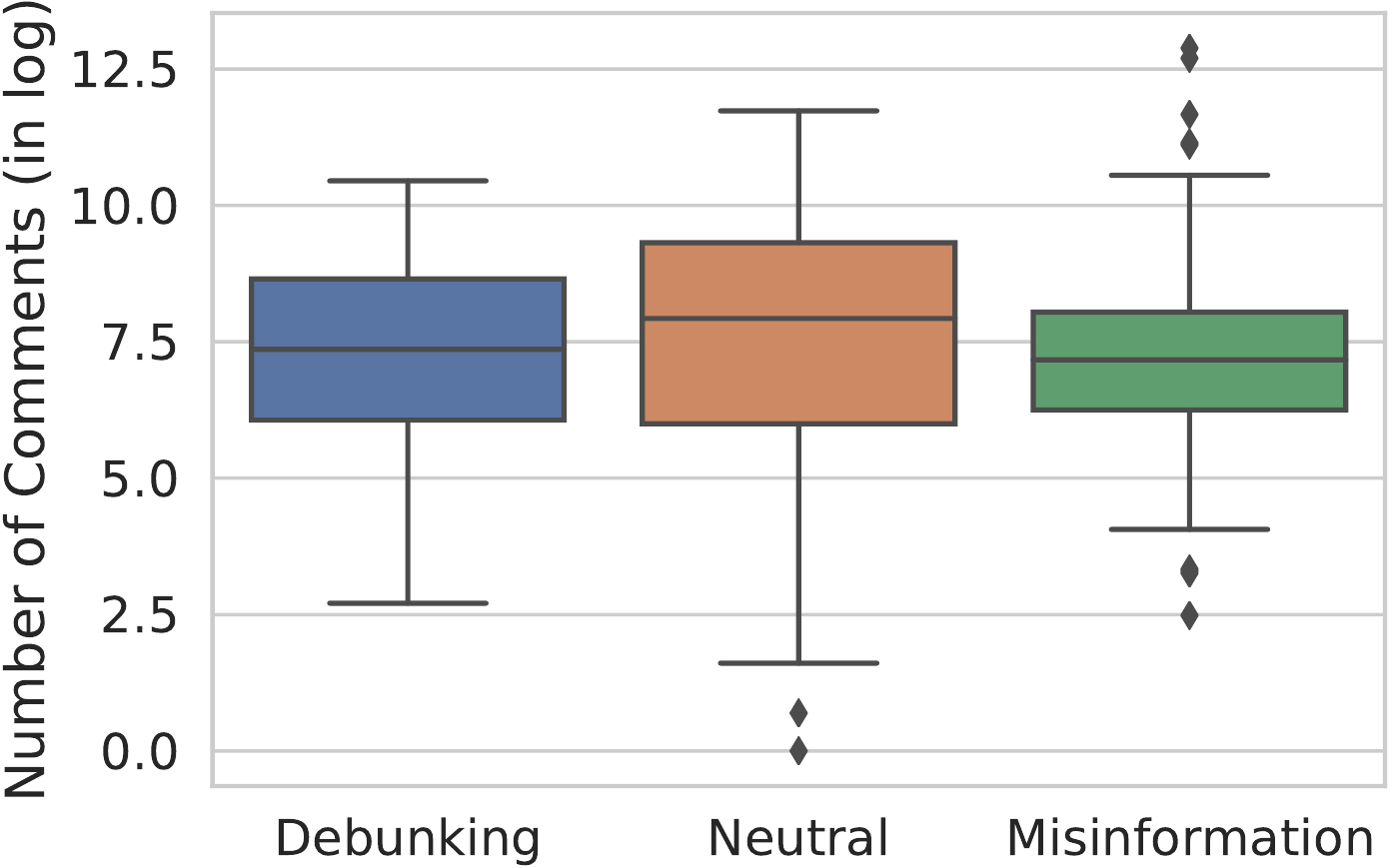}}\\\hline

\rotatebox{90}{\hspace{5mm}Chemtrail} & 
\subfloat{\includegraphics[width=0.45\columnwidth]{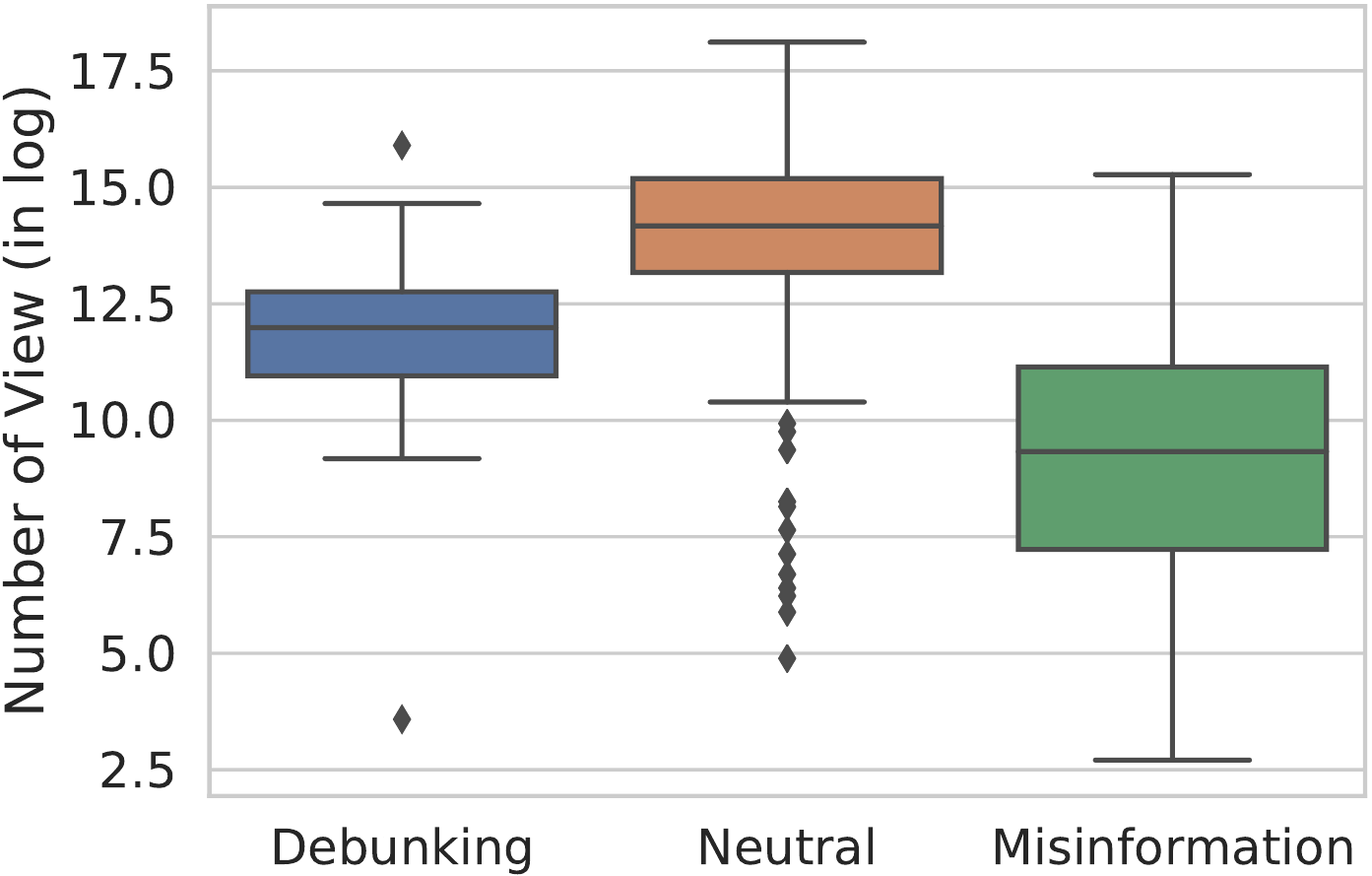}} &
\subfloat{\includegraphics[width=0.45\columnwidth]{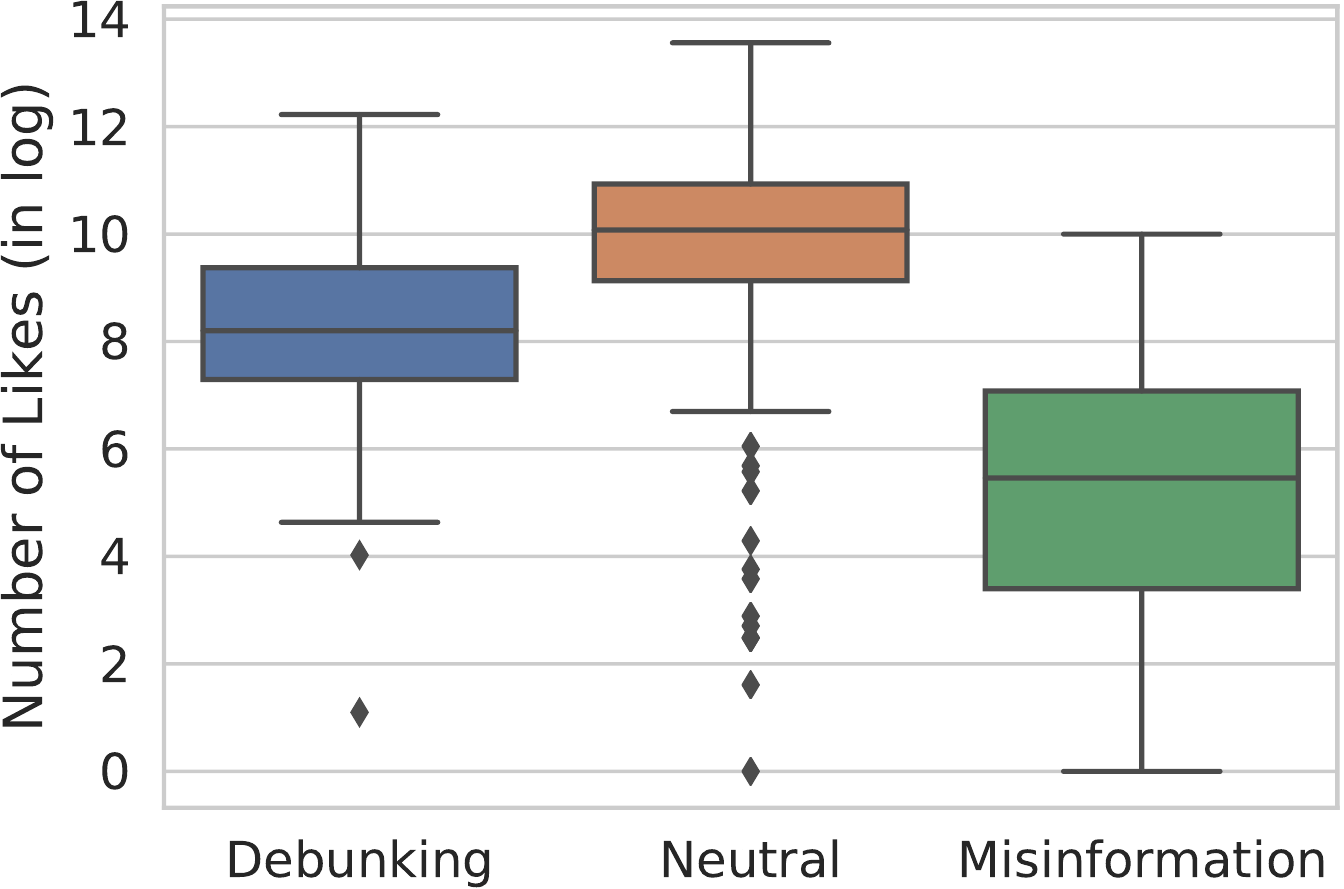}} &
\subfloat{\includegraphics[width=0.45\columnwidth]{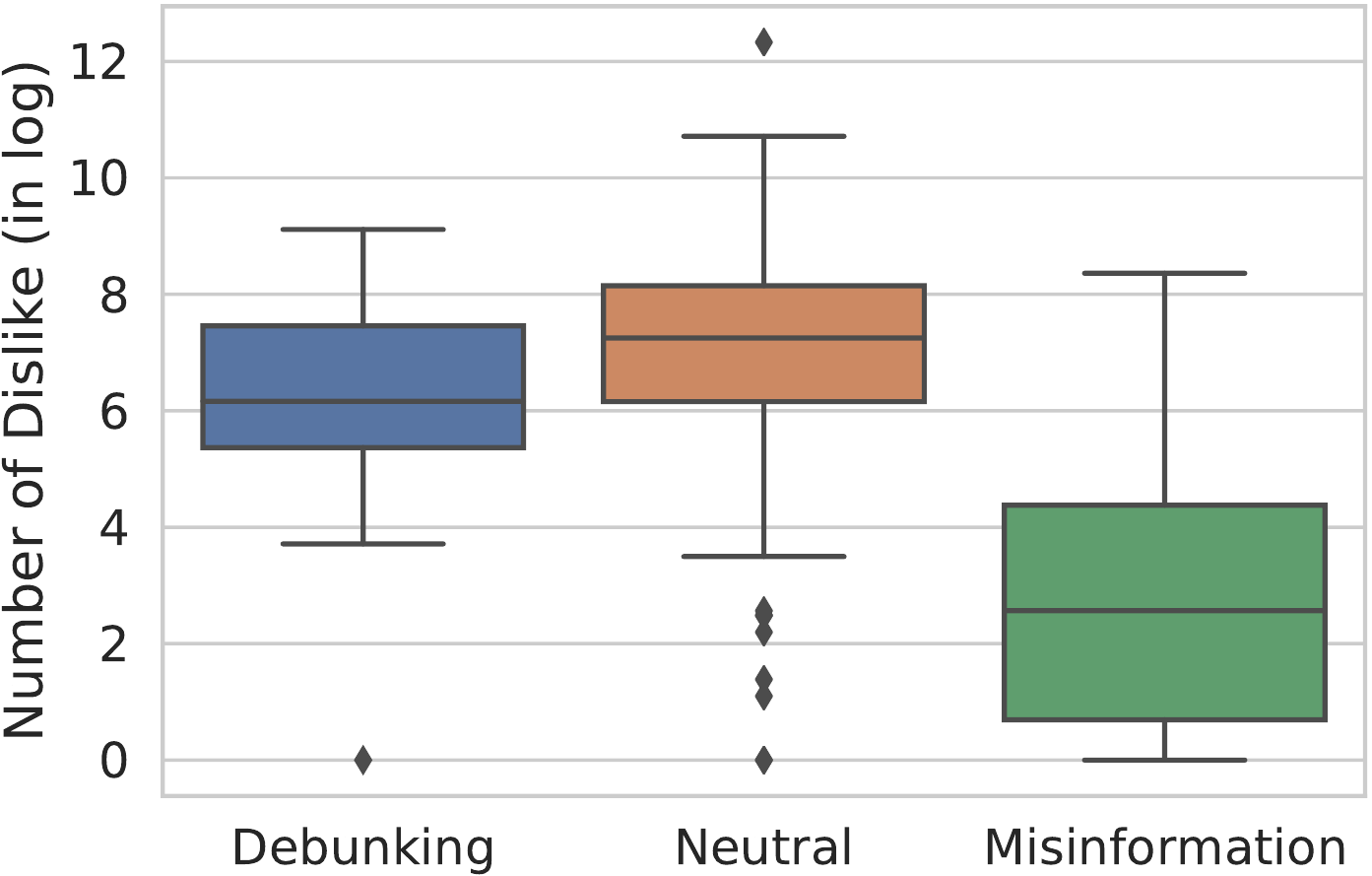}} &
\subfloat{\includegraphics[width=0.45\columnwidth]{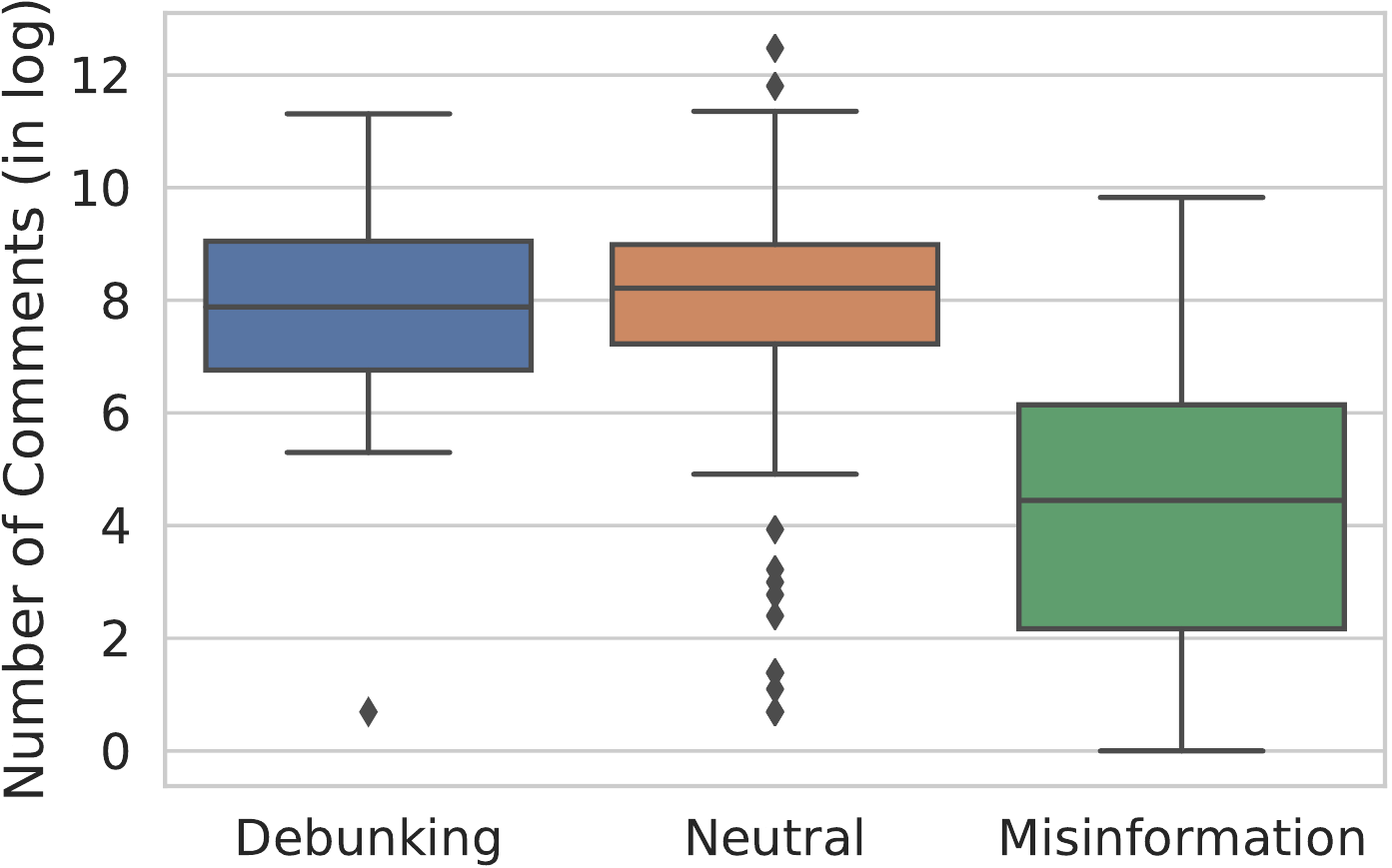}}\\\hline

\rotatebox{90}{\hspace{2mm}Moon Landing} & 
\subfloat{\includegraphics[width=0.45\columnwidth]{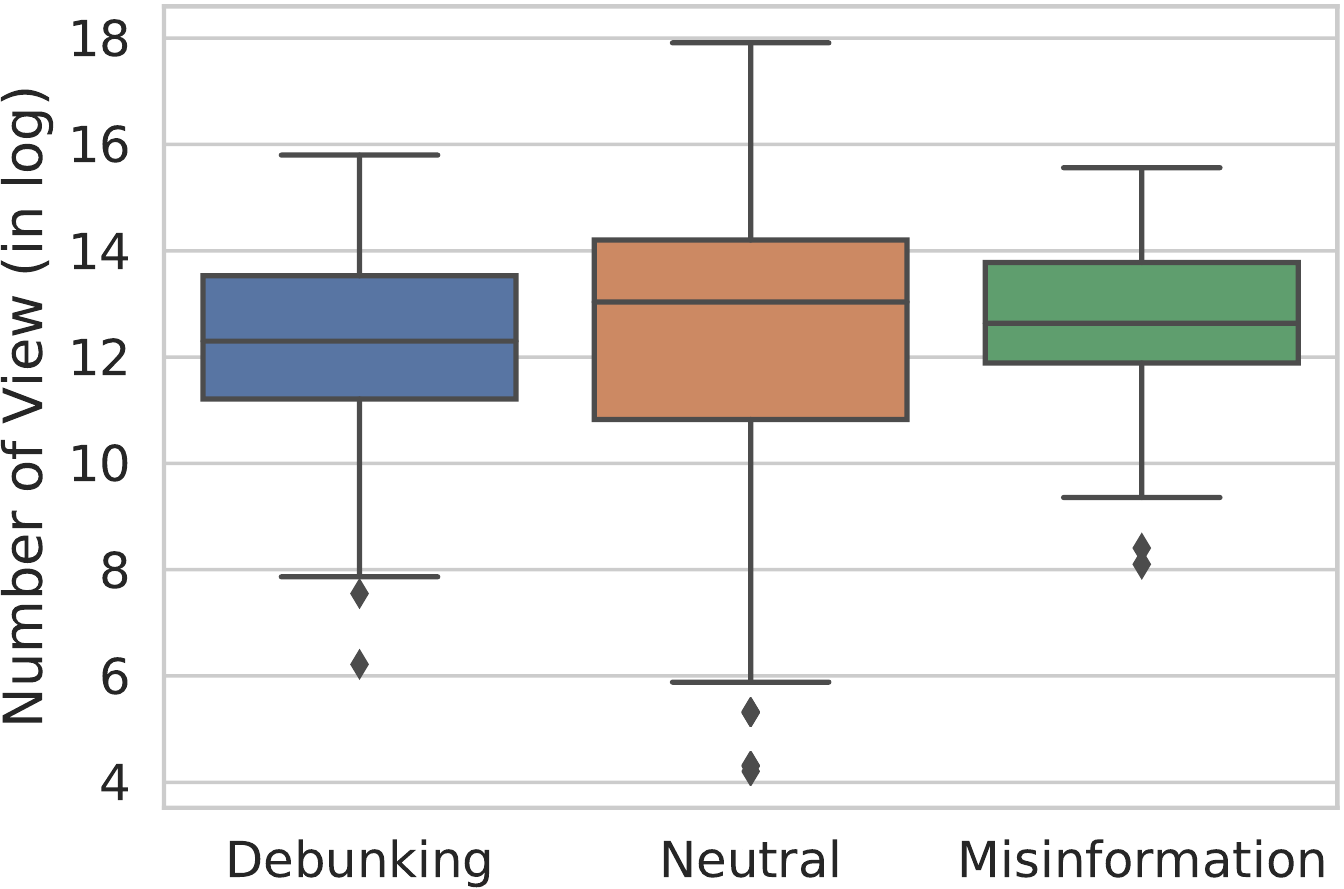}} &
\subfloat{\includegraphics[width=0.45\columnwidth]{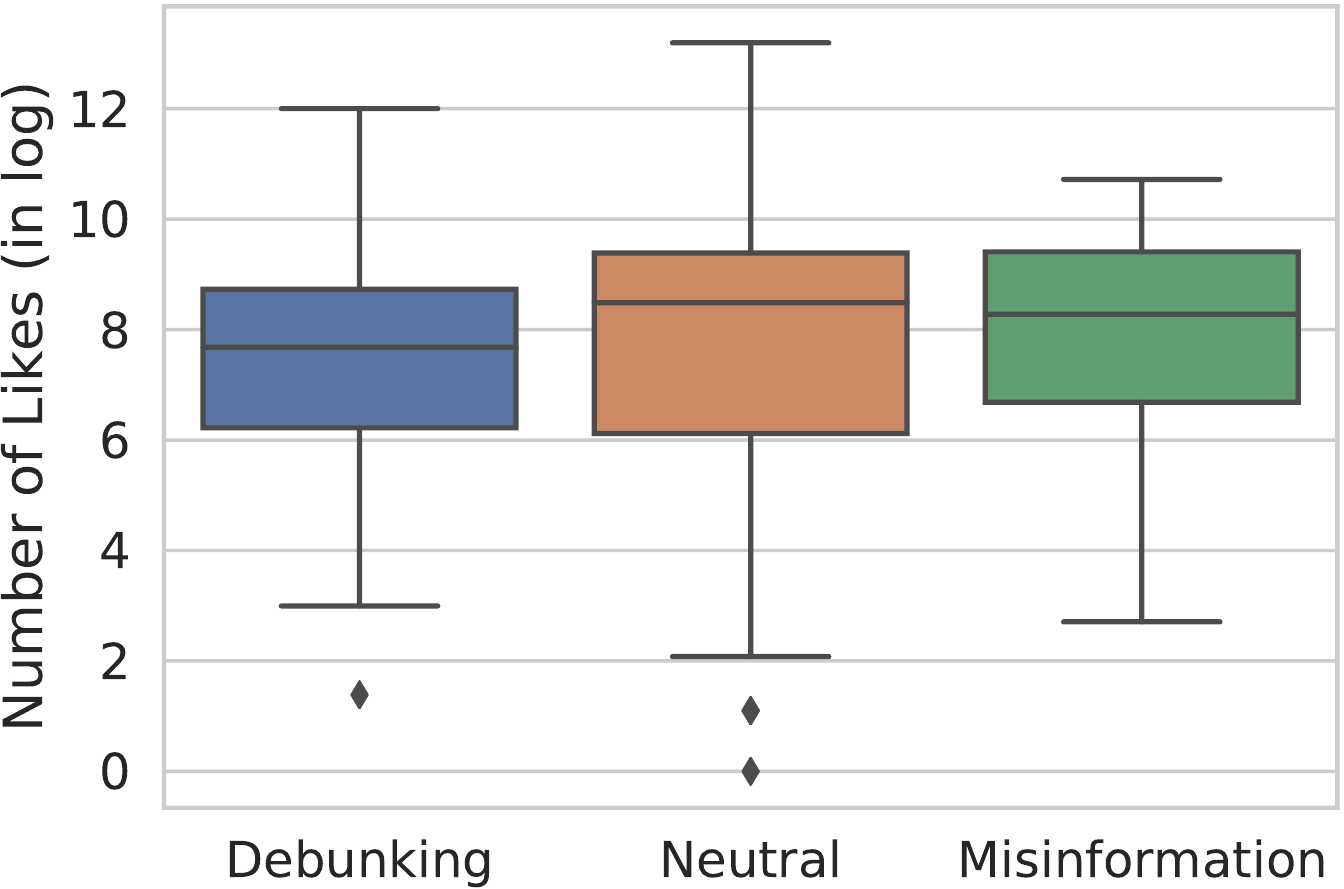}} &
\subfloat{\includegraphics[width=0.45\columnwidth]{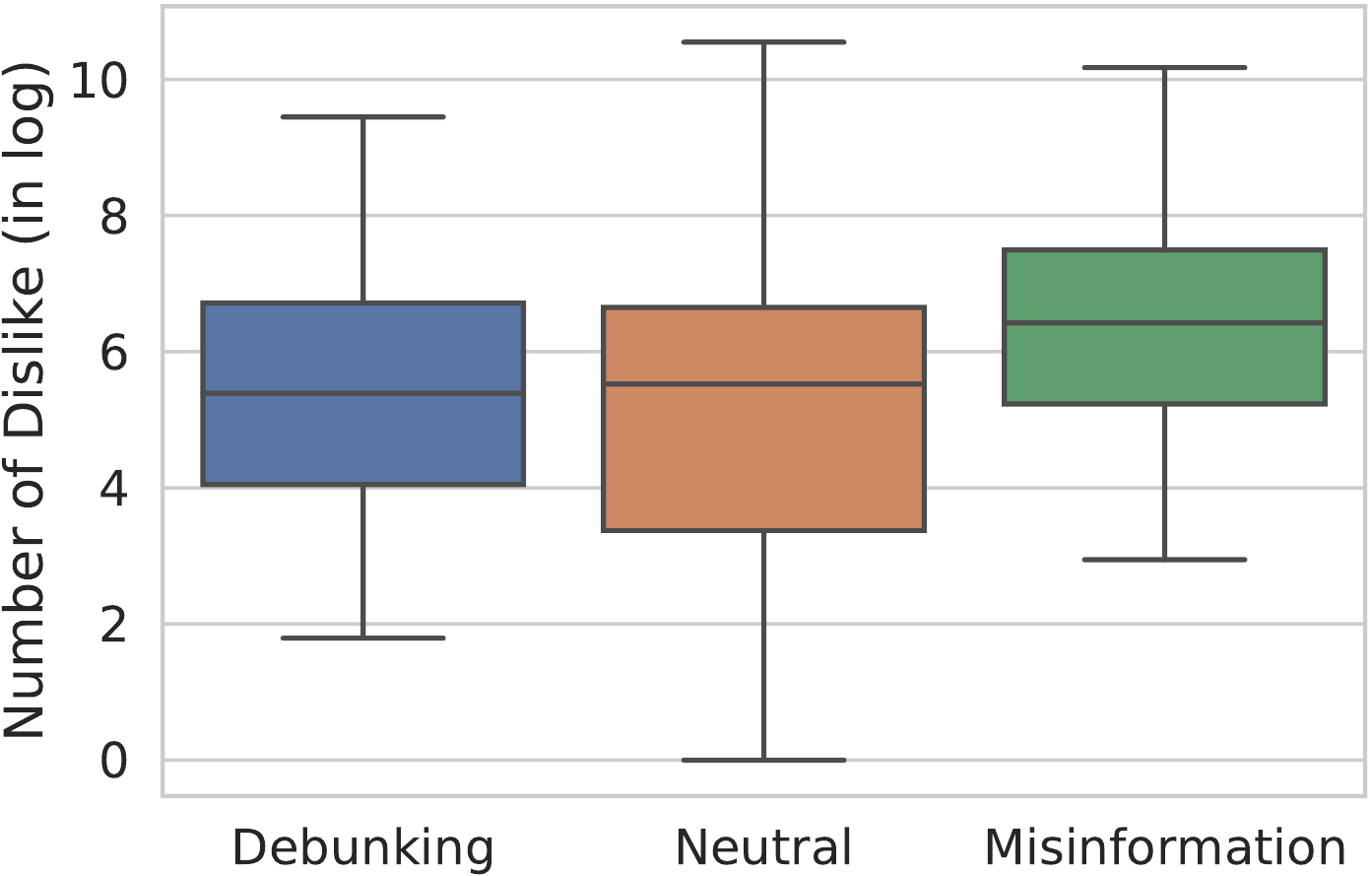}} &
\subfloat{\includegraphics[width=0.45\columnwidth]{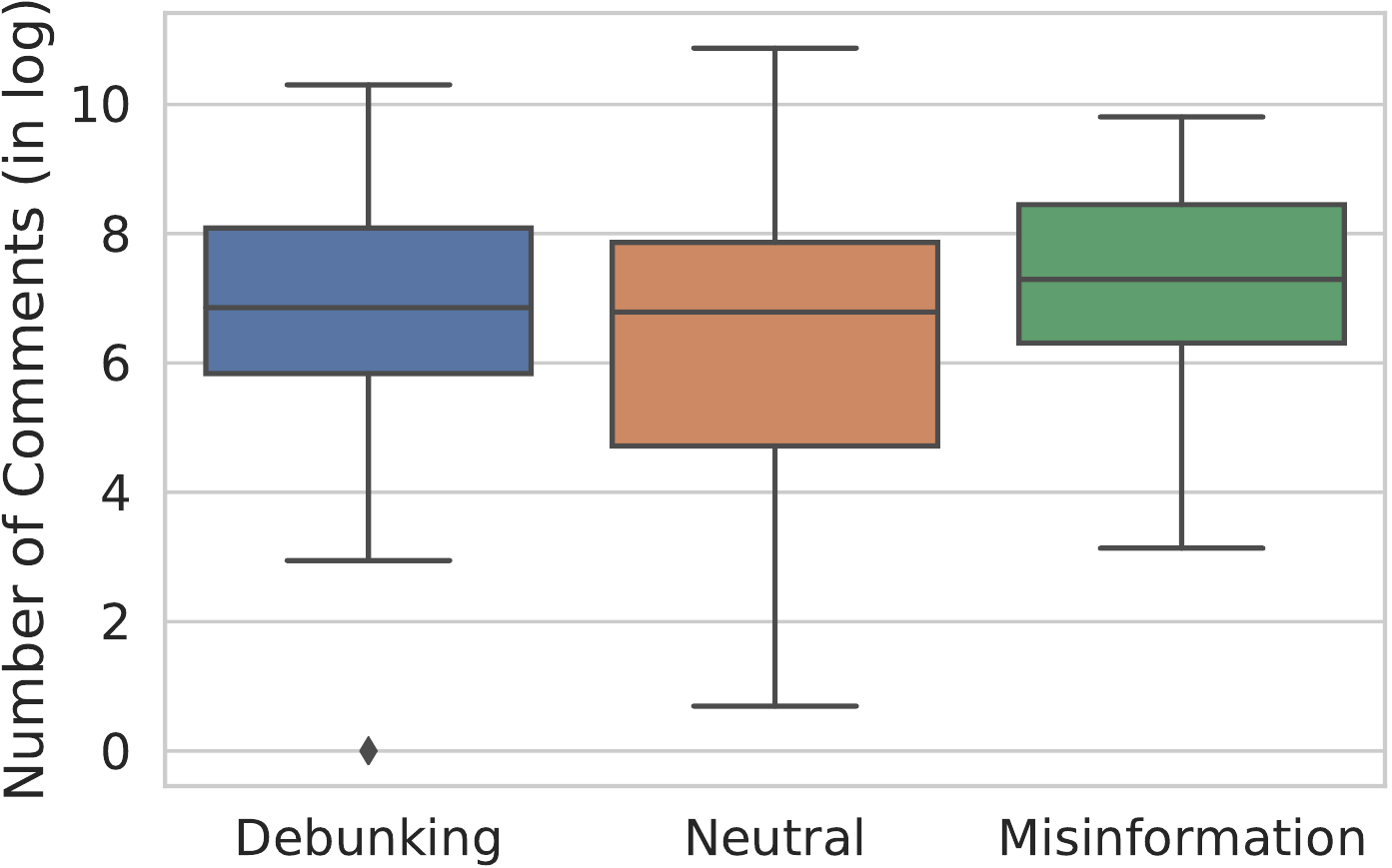}}\\\hline

\rotatebox{90}{\hspace{5mm}Flat Earth} & 
\subfloat{\includegraphics[width=0.45\columnwidth]{img/Vaccine_Views.pdf}} &
\subfloat{\includegraphics[width=0.45\columnwidth]{img/Vaccine_Like.pdf}} &
\subfloat{\includegraphics[width=0.45\columnwidth]{img/Vaccine_Dislike.pdf}} &
\subfloat{\includegraphics[width=0.45\columnwidth]{img/Vaccine_Comment.pdf}}
\\\hline
\end{tabular}
\caption{
Boxplot statistics of views, likes, dislikes, and comments of Youtube videos belong to topics Vaccine Controversy, 9/11 Conspiracy, Chemtrail Conspiracy, Moon landing Controversy, and Flat Earth.}
\label{fig:BoxPlots}
\end{figure*}

\begin{table*}[h!]
\centering
\begin{tabular}{|p{3cm}|c|c|c|c|c|c|c|c|c|c|c|c|c|c|c|}
\hline
                                      & \multicolumn{3}{|c|}{Vaccines Controversy}            & \multicolumn{3}{|c|}{9/11 Conspiracy}        & \multicolumn{3}{|c|}{Chem-trail Conspiracy}    & \multicolumn{3}{|c|}{Moon Landing Conspiracy}  & \multicolumn{3}{|c|}{Flat Earth}                \\\hline
                                               & M   & DM  & N   & M   & DM  & N   & M   & DM  & N   & M   & DM & N   & M    & DM  & N  \\\hline
Original Videos                                & 87 & 215 & 473  & 65 & 67  &  522 & 52 & 237 & 386  & 28 & 94 & 344  & 33  & 71  &  269    \\\hline
Videos with English caption                    & 47 & 170 & 404  & 49 & 51  &  336 & 140 & 44  & 300 & 21 & 70 & 226  & 22  & 63  &   232               \\\hline
Videos with captions after filtering           & 47 & 160 & 398  & 49 & 50  & 329  & 135 & 44  & 297 & 21 & 66 & 215  & 22  & 62  & 230  \\\hline
Approx English caption length (in thousand)    & 28k & 7k  & 12k & 18k & 14k & 16k & 18k & 10k & 16k & 10k & 9k & 12k & 35k  & 11k &   17k              \\\hline
\end{tabular}
\caption{Preprocessed dataset statistics. Here, M represents Misinformation, DM: Debunking Misinformation, and N: Neutral.}
\label{table:datasetStats}
\end{table*}

In Figure \ref{fig:BoxPlots}, we used boxplots to show the distribution of the number of views, likes, dislikes and comments for each subject and class separately. We notice a common tendency in which neutral videos have more views, likes, and dislikes. One probable explanation is that neutral videos, which are comparatively more prevalent in numbers, are more likely to be seen, resulting in a higher number of views, likes, and dislikes. In Figure \ref{fig:BoxPlots} (Row 4) for the Chem-trail Conspiracy topic, we observe an intriguing fact that the number of comments on misinformation videos is more than the number of comments on neutral videos. Furthermore, the number of views, likes, and dislikes for the chem-trail topic are more comparable across all the three classes - neutral, misinformation, and debunking misinformation. We also notice that the misinformation and debunking misinformation videos for all topics follow a similar pattern and are hard to differentiate. This motivates us to collect the video's caption to make a better distinction among the classes. After retrieving the captions using the Caption Scraper, we pre-processed them as already mentioned in Section \ref{subsec:captionCollPreprocess}. In Table \ref{table:datasetStats}, we report caption dataset information after preprocessing for each topic and all classes. We determined the average length of the captions as shown in Table \ref{table:datasetStats} (last row). There is no general pattern in terms of the length of the captions. Thus, we utilized the actual text in the captions for our predictive model for predicting misinformation videos.

\section{Proposed Methodology}\label{sec:methodology}

\begin{figure*}[h!]
    \includegraphics[width=\linewidth]{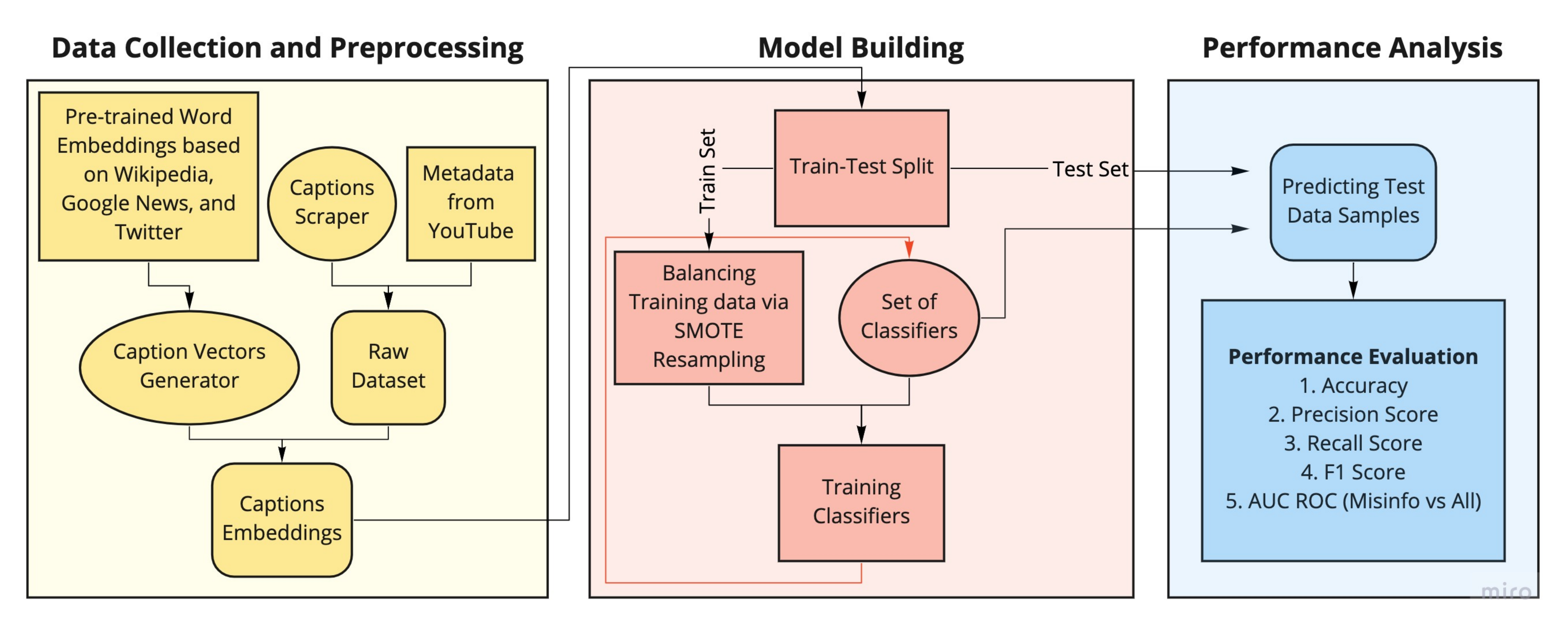}%
    \caption{Pipeline of the proposed methodology. See details in Section \ref{sec:methodology}.}
    \label{fig:methodology_pipeline}
\end{figure*}

In this section, we describe our methodology (see Figure \ref{fig:methodology_pipeline}) in detail.

\subsection{Data Collection and Preprocessing}
The dataset \cite{YTdataset} consists of YouTube videos links containing information such as the number of likes, number of views. The dataset is categorized into three classes, Misinformation (1), Debunking Misinformation (-1), and Neutral (0). In addition, we also collected the captions of each video using the links present in the dataset using Caption Scraper (see Section \ref{Sec:Dataset} for more details. Figure \ref{fig:methodology_pipeline} left box shows this part).


After collecting the dataset, we transform the preprocessed video-captions text into a numerical representation vector. We employed the following four state-of-the-art pre-trained word-to-vector embeddings on video captions to generate numerical vectors \cite{mikolov2013efficient} \cite{ pennington2014glove}. 
\begin{enumerate}
    \item Stanford GloVe Wikipedia vectors - $100D$
    \item Stanford GloVe Wikipedia vectors - $300D$
    \item Word2Vec Google News - $300D$
    \item Word2Vec Twitter - $200D$
\end{enumerate}

In GloVe, the distance between vectors in the respective embedding space is proportional to the relative co-occurrence of words they represent. That means if two words co-occur in multiple articles, they are likely to be similar and close in the vector space. Word2Vec is a two-layer neural network approach that tries to learn the hidden layer weights to predict the vector embedding for words. We generate finite-dimensional caption vectors or embeddings based on these approaches. Each caption vector is formulated as the weighted average of the vector representations of each word in the caption.  

\subsection{Model Building}

As we already discussed in Section \ref{Sec:Dataset} that our dataset has three different classes. Therefore, our problem is a multi-class video classification problem. 
We assume we have a training data $(\bx_i, y_i), i = 1, 2, \ldots, N$, where $N$ represents the number of data samples. Here each $\bx_i \in \Real^D$ is $D$-dimensional feature vector, which we obtain by various embedding schemes on each video caption. Each Youtube video is labeled to one of the three classes, i.e.,  $y_i \in \{-1, 0, 1 \}$. Given the training data, we aim to learn a classifier that can label an unseen video represented in the same training data space into the three classes $\{-1, 0, 1\}$.

We are also interested in identifying misinformation videos in the cloud video-sharing platforms to filter them. To this end, we consider videos labeled debunking misinformation and neutral videos as one class (Class 0) and the misinformation videos as the other class (Class 1), posing this a binary-class classification problem.

From Figure \ref{fig:my_label}, we observe that, on average, videos belonging to different classes are imbalanced for all topics---i.e. a \textit{imbalanced multi-class classification} \cite{zhang2015towards} problem. We thus use the popular Synthetic Minority Over-sampling Technique (SMOTE) \cite{chawla2002smote} to resolve the class imbalance problem in the training data. SMOTE works by selecting examples close in the feature space, drawing a line between the examples in the feature space, and drawing a new sample at a point along that line. Specifically, a random example $\bx_i$ from the minority class is chosen first, and its $k$-nearest neighbors are found (typical $k = 5$). A neighbor $\bx_j$ is chosen randomly, and finally, a synthetic example is created at a randomly selected point between the two examples $(\bx_i, \bx_j)$ in the feature space.

\subsection{Analysing the Performance}

We calculate the F1 scores, AUC-ROC, Precision Score, and Recall Score to identify the best classifiers with the best embeddings for all the topics. (See Section \ref{sec:experiments} for more details. Figure ~\ref{fig:methodology_pipeline} right most box shows this part).

\section{Experimental Analysis}\label{sec:experiments}
Next, using caption-text vectorization explained in the previous section, we build predictive models to estimate the likelihood that a video belongs to a specific class.

\subsection{Experimental setup}\label{sec:eval_metrics}

We use a set of classifiers from the Python library LazyPredict\footnote{https://pypi.org/project/lazypredict/} that contains 28 classifiers such as Support Vector Classifier (SVC), $k$-neighbor classifier ($k$-nn), Random Forests,  and XGBoost. To evaluate the performance of these classifiers, we used standard metrics such as (i) F$1$-score (weighted averaging), (ii) AUC-ROC, (iii) Precision (weighted averaging), (iv) Recall (weighted averaging), and (v) Accuracy. For our multi-class classification problem, the formula to calculate these matrices are as follows. 

Let $N$ be the total number of data points to be classified into $K$ classes, and $n_k$ be the number of true labels (data points) for class $k$.  
\begin{eqnarray}
\text{Precision}_k &=& \frac{TP_k}{TP_k + FP_k} \\ 
\text{Precision}_\text{weighted} &=& \frac{1}{N} \sum_{k = 1}^K n_k \: \text{Precision}_k \\ 
\text{Recall}_k &=& \frac{TP_k}{TP_k + FN_k} \\ 
\text{Recall}_\text{weighted} &=& \frac{1}{N} \sum_{k = 1}^K n_k \: \text{Recall}_k\\
\text{F$1$-score}_k &=& 2*\frac{\text{Precision}_{k} * \text{Recall}_{k}}{\text{Precision}_{k} + \text{Recall}_{k}}\\
\text{F$1$-score}_\text{weighted} &=& \frac{1}{N} \sum_{k = 1}^K n_k \: \text{F$1$-score}_k
\end{eqnarray}


As mentioned earlier in Section \ref{sec:methodology} that we use four different embedding (text vectorization) schemes. Therefore, we defined a scoring method to calculate \textbf{embedding performance} given the top-\emph{T} number of classifiers. Note that we rank all the models based on their F$1$-score, which means the model with a higher F$1$-score is better than a model with a lower F$1$-score. For each embedding, embedding performance can be calculated using the mean of top-\emph{T} classifiers' F$1$-score as:
\begin{equation}
    \mu_{\text{embed}(j)} = \frac{1}{T} \sum_{k = 1}^{T} \text{F$1$-score}_{\text{embed}(j), k} 
    \label{eq:scoring}
\end{equation}
where $j$ represents the $j^{th}$ embedding (or text vectorization) scheme. The purpose of using this scoring method is two-fold: (a) It uses best representative classifiers for each embedding, e.g., if SVC is best for Embedding 1, then it will be compared with the best classifier of Embedding 2 that may not be SVC, and (b) It handles high scores outliers by averaging F$1$-score of top-\emph{T} classifiers. 

\subsection{Results}\label{sec:all_class_classifier}
Table \ref{table:3class_best_classifiers} shows the best classifiers and embedding based on F$1$-score for each topic. Please note that a Dummy Classifier will give $0.35$ F$1$-score; however, using caption vectors and our best classifier, we able to classify videos in three classes,  Misinformation (1), Debunking Misinformation (-1), Neutral (1) with $0.85 \text{ to } 0.90$ F$1$-score. Next, using scoring method discussed before, we computed the embedding performance with three different number of top-$T$ models ($T = 5, 10,15$) for each topic (see Table \ref{table:3class_emb_score}). We observe that the embedding trained on Google News is the best for representing the misinformation-prone caption data.



\begin{table}[]
\centering
\begin{tabular}{|l|r|r|r|r|}
\hline
\multicolumn{5}{|c|}{Three-class Classifier (sorted by F$1$-score)}                                                                                                                                                                          \\ \hline
\multicolumn{1}{|c|}{Topics}                                                                              & \multicolumn{1}{c|}{F$1$-score} & \multicolumn{1}{c|}{Precision} & \multicolumn{1}{c|}{Recall} & \multicolumn{1}{c|}{Accuracy} \\ \hline
\begin{tabular}[c]{@{}l@{}}\textsl{Vaccines Controversy}\\ NuSVC\\ Google 300D\end{tabular}                        & 0.89                          & 0.89                           & 0.89                        & 0.89                          \\ \hline
\begin{tabular}[c]{@{}l@{}}\textsl{9/11 Conspiracy}\\ ExtraTreesClassifier\\ Twitter 200D\end{tabular}             & 0.90                          & 0.91                           & 0.91                        & 0.91                          \\ \hline
\begin{tabular}[c]{@{}l@{}}\textsl{Chem-trail Conspiracy}\\ CalibratedClassifierCV\\ Google News 300D\end{tabular} & 0.89                          & 0.89                           & 0.89                        & 0.89                          \\ \hline
\begin{tabular}[c]{@{}l@{}}\textsl{Flat Earth}\\ RandomForestClassifier\\ GloVe 300D\end{tabular}                  & 0.86                          & 0.87                           & 0.85                        & 0.85                          \\ \hline
\begin{tabular}[c]{@{}l@{}}\textsl{Moon Landing}\\ NuSVC\\ GloVe 300D\end{tabular}                                 & 0.85                          & 0.84                           & 0.85                        & 0.85                          \\ \hline
\end{tabular}
\caption{\textbf{Multi-class classification: }Best models and embeddings with highest weighted F$1$-score for each topic.}
\label{table:3class_best_classifiers}
\end{table}



\begin{table}[]
\centering
\begin{tabular}{|l|c|r|r|r|r|}
\hline
\multicolumn{1}{|c|}{Topics}           & \multicolumn{1}{c|}{T} & \multicolumn{1}{c|}{\begin{tabular}[c]{@{}c@{}}GloVe\\ 100D\end{tabular}} & \multicolumn{1}{c|}{\begin{tabular}[c]{@{}c@{}}GloVe\\ 300D\end{tabular}} & \multicolumn{1}{c|}{\begin{tabular}[c]{@{}c@{}}Google\\ News\\ 300D\end{tabular}} & \multicolumn{1}{c|}{\begin{tabular}[c]{@{}c@{}}Twitter\\ 200D\end{tabular}} \\ \hline
\multirow{3}{*}{Vaccines Controversy}  & 5                 & 0.85                                                                      & 0.87                                                                      & 0.87                                                                              & 0.86                                                                        \\ \cline{2-6} 
                                       & 10                & 0.83                                                                      & 0.84                                                                      & 0.85                                                                              & 0.86                                                                        \\ \cline{2-6} 
                                       & 15                & 0.82                                                                      & 0.82                                                                      & 0.84                                                                              & 0.85                                                                        \\ \hline
\multirow{3}{*}{9/11 Conspiracy}       & 5                 & 0.82                                                                      & 0.85                                                                      & 0.85                                                                              & 0.84                                                                        \\ \cline{2-6} 
                                       & 10                & 0.79                                                                      & 0.81                                                                      & 0.83                                                                              & 0.81                                                                        \\ \cline{2-6} 
                                       & 15                & 0.78                                                                      & 0.79                                                                      & 0.81                                                                              & 0.80                                                                        \\ \hline
\multirow{3}{*}{Chem-trail Conspiracy} & 5                 & 0.82                                                                      & 0.84                                                                      & 0.88                                                                              & 0.84                                                                        \\ \cline{2-6} 
                                       & 10                & 0.80                                                                      & 0.82                                                                      & 0.87                                                                              & 0.82                                                                        \\ \cline{2-6} 
                                       & 15                & 0.79                                                                      & 0.80                                                                      & 0.84                                                                              & 0.81                                                                        \\ \hline
\multirow{3}{*}{Flat Earth}            & 5                 & 0.75                                                                      & 0.80                                                                      & 0.76                                                                              & 0.76                                                                        \\ \cline{2-6} 
                                       & 10                & 0.72                                                                      & 0.78                                                                      & 0.74                                                                              & 0.75                                                                        \\ \cline{2-6} 
                                       & 15                & 0.70                                                                      & 0.75                                                                      & 0.72                                                                              & 0.73                                                                        \\ \hline
\multirow{3}{*}{Moon Landing}          &  5                 & 0.80                                                                      & 0.81                                                                      & 0.80                                                                              & 0.80                                                                        \\ \cline{2-6} 
                                       & 10                & 0.76                                                                      & 0.78                                                                      & 0.76                                                                              & 0.77                                                                        \\ \cline{2-6} 
                                       & 15                & 0.73                                                                      & 0.75                                                                      & 0.73                                                                              & 0.75                                                                        \\ \hline
\end{tabular}
\caption{Embedding performance using scoring method \eqref{eq:scoring} for multi-class classification.}
\label{table:3class_emb_score}
\end{table}

\paragraph*{Misinformation classifier} This work focuses on the identification of misinformation. As a result, to emphasize the relevance of the misinformation class, we re-formulate our classification problem as a two-class classification - Misinformation and others (Neutral and Debunking Misinformation). Table \ref{table:2class_best_classifiers} shows the best classifiers and embedding based on F$1$-score for each topic. Using caption vectors and our best classifier, we can identify misinformation videos with $0.92 \text{ to } 0.95$ F$1$-score and $0.78 \text{ to } 0.90$ AUC ROC. Next, using scoring method, we computed the embedding performance with three different number of top-$T$ models ($T = 5, 10,15$) for each topic (see Table \ref{table:2class_emb_score}). We can observe that Google News $300D$ and GloVe $300D$ are the best representations of video captions of the above topics. We can also observe that embeddings trained on the Twitter dataset have a better representation for a political topic such as the 9/11 conspiracy.

\begin{table}[]
\centering
\begin{tabular}{|p{2.35cm}|r|r|r|r|r|}
\hline
\multicolumn{6}{|c|}{Misinformation Classifier (sorted by F$1$-score)}                                                                                                                                                                                       \\ \hline
\multicolumn{1}{|c|}{Topics}                                                                & \multicolumn{1}{c|}{F1} & \multicolumn{1}{c|}{Precision} & \multicolumn{1}{c|}{Recall} & \multicolumn{1}{c|}{Accuracy} & \multicolumn{1}{l|}{AUC} \\
\multicolumn{1}{|c|}{}                                                                & \multicolumn{1}{c|}{Score} & \multicolumn{1}{c|}{} & \multicolumn{1}{c|}{} & \multicolumn{1}{c|}{} & \multicolumn{1}{l|}{ROC} \\ \hline
\begin{tabular}[c]{@{}l@{}}\textsl{Vaccines Controversy}\\ SVC\\ GloVe 100D\end{tabular}             & 0.97                          & 0.97                           & 0.97                        & 0.97                          & 0.89                          \\ \hline
\begin{tabular}[c]{@{}l@{}}\textsl{9/11 Conspiracy}\\ ExtraTreesClassifier\\ GloVe 300D\end{tabular} & 0.93                          & 0.93                           & 0.94                        & 0.94                          & 0.78                          \\ \hline
\begin{tabular}[c]{@{}l@{}}\textsl{Chem-trail Conspiracy}\\ XGBClassifier\\ GloVe 300D\end{tabular}  & 0.92                          & 0.92                           & 0.92                        & 0.92                          & 0.90                          \\ \hline
\begin{tabular}[c]{@{}l@{}}\textsl{Flat Earth}\\ AdaBoostClassifier\\ Google 300D\end{tabular}       & 0.96                          & 0.96                           & 0.96                        & 0.96                          & 0.82                          \\ \hline
\begin{tabular}[c]{@{}l@{}}\textsl{Moon Landing}\\ SVC\\ Google 300D\end{tabular}                    & 0.96                          & 0.96                           & 0.96                        & 0.96                          & 0.74                          \\ \hline
\end{tabular}
\caption{\textbf{Binary-class classification: }Best models and embeddings with highest weighted F$1$-score for each topic.}
\label{table:2class_best_classifiers}
\end{table}

\begin{table}[]
\centering
\begin{tabular}{|l|c|r|r|r|r|}
\hline
\multicolumn{1}{|c|}{Topics}           & \multicolumn{1}{c|}{$T$} & \multicolumn{1}{c|}{\begin{tabular}[c]{@{}c@{}}GloVe\\ 100D\end{tabular}} & \multicolumn{1}{c|}{\begin{tabular}[c]{@{}c@{}}GloVe\\ 300D\end{tabular}} & \multicolumn{1}{c|}{\begin{tabular}[c]{@{}c@{}}Google\\ News\\ 300D\end{tabular}} & \multicolumn{1}{c|}{\begin{tabular}[c]{@{}c@{}}Twitter\\ 200D\end{tabular}} \\ \hline
\multirow{3}{*}{Vaccines Controversy}  & 5                 & 0.95                                                                      & 0.96                                                                      & 0.96                                                                              & 0.96                                                                        \\ \cline{2-6} 
                                       & 10                & 0.94                                                                      & 0.95                                                                      & 0.95                                                                              & 0.95                                                                        \\ \cline{2-6} 
                                       & 15                & 0.93                                                                      & 0.93                                                                      & 0.95                                                                              & 0.94                                                                        \\ \hline
\multirow{3}{*}{9/11 Conspiracy}       & 5                 & 0.91                                                                      & 0.91                                                                      & 0.91                                                                              & 0.93                                                                        \\ \cline{2-6} 
                                       & 10                & 0.89                                                                      & 0.90                                                                      & 0.90                                                                              & 0.91                                                                        \\ \cline{2-6} 
                                       & 15                & 0.89                                                                      & 0.89                                                                      & 0.89                                                                              & 0.90                                                                        \\ \hline
\multirow{3}{*}{Chem-trail Conspiracy} & 5                 & 0.87                                                                      & 0.90                                                                      & 0.89                                                                              & 0.88                                                                        \\ \cline{2-6} 
                                       & 10                & 0.87                                                                      & 0.88                                                                      & 0.88                                                                              & 0.87                                                                        \\ \cline{2-6} 
                                       & 15                & 0.86                                                                      & 0.87                                                                      & 0.87                                                                              & 0.86                                                                        \\ \hline
\multirow{3}{*}{Flat Earth}            & 5                 & 0.90                                                                      & 0.91                                                                      & 0.94                                                                              & 0.95                                                                        \\ \cline{2-6} 
                                       & 10                & 0.88                                                                      & 0.90                                                                      & 0.93                                                                              & 0.93                                                                        \\ \cline{2-6} 
                                       & 15                & 0.87                                                                      & 0.89                                                                      & 0.91                                                                              & 0.91                                                                        \\ \hline
\multirow{3}{*}{Moon Landing}          & 5                 & 0.94                                                                      & 0.95                                                                      & 0.95                                                                              & 0.94                                                                        \\ \cline{2-6} 
                                       & 10                & 0.93                                                                      & 0.94                                                                      & 0.94                                                                              & 0.94                                                                        \\ \cline{2-6} 
                                       & 15                & 0.92                                                                      & 0.94                                                                      & 0.94                                                                              & 0.93                                                                        \\ \hline
\end{tabular}
\caption{Embedding performance using scoring method \eqref{eq:scoring} for binary-class classification.}
\label{table:2class_emb_score}
\end{table}

\section{Discussion and Conclusion}\label{Sec:Conclusion}
With a quest to identify misinformation videos on YouTube, we study the YouTube videos dataset of five different topics (Vaccines Controversy, 9/11 Conspiracy, Chem-trail Conspiracy, Moon Landing Conspiracy, and Flat Earth) using various natural language processing techniques. Under each topic, videos are further divided into three classes: Misinformation, Debunking Misinformation, and Neutral. In this work, we exploited the YouTube captions to understand the content of the videos using multiple pre-trained word embeddings to convert video captions to D-dimensional vector space. Our analysis showed that we could classify videos among three classes with $0.85 \text{ to } 0.90$ F$1$-score. Furthermore, to emphasize the relevance of the misinformation class, we re-formulate our three-class problem as a two-class classification - Misinformation and others (Debunking Misinformation and Neutral). In this case, our model could classify videos with $0.92 \text{ to } 0.95$ F$1$-score and $0.78 \text{ to } 0.90$ AUC ROC.

\paragraph*{Limitations and Future Work} This work has some limitations, primarily related to caption embedding. We used embeddings trained with Wikipedia articles, Google News, and Twitter for the word to vector representations. We selected this set because it includes various data starting from articles open for all platforms Wikipedia to News articles written by professional journalists, from micro-blogging platforms with 280 character limit to huge articles with thousands of words. However, there are other possible techniques to improve these embedding. Future work could consider a new embedding based on YouTube captions that can be trained using the GloVe algorithm, which might work best for YouTube video classification compared to pre-trained embedding. 

\section*{Acknowledgment}
This research is funded by ERDF via the IT Academy Research Programme and H2020 Project, SoBigData++, and CHIST-ERA project, SAI.
\vspace{-3mm}
\bibliographystyle{IEEEtran}
\bibliography{btp}

\end{document}